\documentclass{article}

\usepackage{microtype}
\usepackage{graphicx}
\usepackage{comment}
\usepackage{balance}
\usepackage{subfigure}
\usepackage{booktabs}

\usepackage[accepted]{icml2025}

\usepackage{amsmath}
\usepackage{amssymb}
\usepackage{mathtools}
\usepackage{amsthm}
\usepackage{soul}
\usepackage{graphicx}
\usepackage{xcolor}
\usepackage{tikz}
\usepackage{fontawesome5}
\usepackage{adjustbox}
\usepackage{multirow}
\usepackage{placeins}
\usepackage{enumitem}
\usepackage{float}

\usepackage[capitalize,noabbrev]{cleveref}

\theoremstyle{plain}

\theoremstyle{definition}

\theoremstyle{remark}

\usepackage[textsize=tiny]{todonotes}

\icmltitlerunning{Hallucination Detection: A Probabilistic Framework Using Embeddings Distance Analysis}

\begin{document}

\twocolumn[
\icmltitle{Hallucination Detection: A Probabilistic Framework Using Embeddings Distance Analysis}

\icmlsetsymbol{equal}{*}

\begin{icmlauthorlist}
\icmlauthor{Emanuele Ricco}{kaust}
\icmlauthor{Lorenzo Cima}{unipi,cnr}
\icmlauthor{Roberto Di Pietro}{kaust}
\end{icmlauthorlist}

\icmlaffiliation{unipi}{University of Pisa, Department of Information Engineering, Pisa, Italy}
\icmlaffiliation{cnr}{IIT-CNR, Pisa, Italy}
\icmlaffiliation{kaust}{King-Abdullah University of Science and Technology (KAUST), CEMSE division, Thuwal, Saudi Arabia}

\icmlcorrespondingauthor{Emanuele Ricco}{emaunele.ricco@kaust.edu.sa}

\icmlkeywords{hallucinations detection, Minkowski distance, embedding, LLMs}

\vskip 0.3in
]

\printAffiliationsAndNotice{}  

\begin{abstract}
Hallucinations are one of the major issues affecting LLMs, hindering their wide adoption in production systems. While current research solutions for detecting hallucinations are mainly based on heuristics, in this paper we introduce a mathematically sound methodology to reason about hallucination, and leverage it to build a tool to detect hallucinations.
To the best of our knowledge, we are the first to show that hallucinated content has structural differences with respect to correct content. To prove this result, we resort to the Minkowski distances in the embedding space. Our findings demonstrate statistically significant differences in the embedding distance distributions, that are also scale free---they qualitatively hold regardless of the distance norm used and the number of keywords, questions, or responses. We leverage these structural differences to develop a tool to detect hallucinated responses, achieving an accuracy of 66\% for a specific  configuration of system parameters---comparable with the best results in the field. 
In conclusion, 
the suggested methodology is promising and novel, possibly paving the way for further research in the domain, also along the directions highlighted in our future work.

\end{abstract}

\section{Introduction} \label{sec:introduction}
Hallucinations in the context of artificial intelligence, particularly in large language models (LLMs), refer to generating content that appears plausible or convincing while it is factually incorrect, misleading, or entirely fabricated. These phenomena 
stem from 
the probabilistic and predictive mechanisms underlying LLMs. Hallucinations can be classified in \textit{intrinsic hallucinations}, when the response has an opposite meaning with respect to the source material and \textit{extrinsic hallucinations}, when the correctness of the response cannot be directly proved ~\cite{huang2023survey}. Hallucinations can occur in various forms, including incorrect facts, non-sensical statements, or invented sentence, resulting from the model’s reliance on statistical patterns to generate responses and could be due to many causes. One possible cause 
is the existence of misleading information in the training data that could lead to produce erroneous answers---a modern version of the thrash-in-thrash-out paradigm \cite{lin2022truthfulqa}. \\
An example of the danger of tainted data is the presence of sociocultural biases based on nationality stereotypes, that associate different entities not supported in the training data ~\cite{ladhak2023pre}. 
The spreading of hallucinations in complex social networks could leverage public opinion and administrative processes from governments, possibly originating social disorders and distrust in AI-generated content \cite{hao2024quantifying}, in particular if incorrect information is spread in a coordinated manner~\cite{cima2024coordinated}. Current detection strategies can be grouped in factuality and faithfulness approaches ~\cite{huang2023survey}. While the first one involves a semiautomated fact-checking approach that needs human participation, ~\cite{chern2023factool}, the second one requires contextual information to analyze the truthness ~\cite{fabbri2021summeval}. 
However, there have been no efforts into investigating the structural distinctions between genuine and hallucinated responses in the context of hallucination detection. 
Our research addresses this gap by introducing a novel perspective: analyzing the embedding space to uncover patterns and discrepancies that differentiate authentic outputs from hallucinated ones. We aim to pave the way for a more robust and nuanced approach to detect hallucinations, offering a fresh direction for both theoretical and practical advancements in this field.

\subsection{Contributions} \label{sub:contributions}
Our contributions can be summarized as follows:
\begin{itemize}
    \item To the best of our knowledge, we are the first to propose a mathematically rigorous method, based on embedding distances, to evaluate the structural differences between genuine and hallucinated responses. 
    \item We demonstrate that, regardless of the distance norm used and the number of keywords, questions, or responses, there is a statistically significant structural difference between genuine and hallucinated LLM responses.
    \item We provide an algorithm to leverage these structural differences to detect whether a response was hallucinated, evaluating its performance using different configurations. Results rank among the top ones, with an accuracy of $66\%$ for a specific configuration of system parameters.
\end{itemize}

\subsection{Roadmap} \label{sec:roadmap}
The remainder of this paper is organized as follows. Section \ref{sec:relworks} discusses previous work related to hallucination detection. 
Section \ref{sec:methodology}  presents our proposed methodology for both training and test data, while Section \ref{sec:results}  reports the results obtained from the model. Section \ref{sec:discussion}  discusses implications and limitations of our proposal, together with some intuitions for future work. Finally, Section \ref{sec:conclusions}  concludes the paper.  \section{Related Work} \label{sec:relworks}
Hallucinations are defined as ``a phenomenon in which the generated content appears non-sensical or unfaithful to the provided source content"~\cite{filippova2020controlled, maynez2020faithfulness}. They affect all LLMs~\cite{huang2023survey}, due to the very architecture of LLMs. Common causes include vague prompts~\cite{ren2022out} and gaps in the training data~\cite{perkovic2024hallucinations}, leading for instance to inaccuracies or perpetuation of societal biases~\cite{ladhak2023pre, venkit2023nationality, giorgi2024human}. The occurrence of hallucinations can significantly undermine the credibility of responses and complicate mitigation and detection efforts~\cite{guerreiro2022looking, huang2025look}.

\subsection{LLMs Hallucination Mitigation} \label{sub:mitigation} 
For what concerns mitigation, several approaches have been proposed. Some methods improve the prompt context by incorporating tags from reliable sources~\cite{feldman2023trapping} or using knowledge graphs~\cite{martino2023knowledge}, while others use neural networks to predict plausible subsequent tokens and avoid unrelated outputs~\cite{verma2023reducing}. A different technique instead exploits self-evaluations, where models assess the validity of their own outputs~\cite{kadavath2022language, ji2023towards}, and external evaluations, where a powerful LLM is used to evaluate content generated by other models~\cite{hao2024quantifying}. In this way, the output is evaluated and corrected if necessary. Finally, LLMs can also use Retrieval Augmented Generation (RAG) to enhance response credibility---that is, accessing external data source for enhancing their knowledge and possibly reducing errors~\cite{fan2024survey,arslan2024survey}.

\subsection{LLMs Hallucination Detection} \label{sub:detection}
Detecting hallucinations in large language models (LLMs) remains a significant challenge, with various techniques proposed to address this issue. Among the simplest approaches, which also serve mitigation purposes, are self-evaluation methods where the model assesses its own outputs~\cite{kadavath2022language, lin2022teaching, manakul2023selfcheckgpt}, or external evaluations performed by more advanced models, such as GPT-3.5 Turbo, to determine whether each output is realistic or fabricated~\cite{friel2023chainpoll}. 
Other strategies involve token-level analysis, such as examining next-token probabilities~\cite{varshney2023stitch}, evaluating uncertainty for sequence of tokens~\cite{malinin2021uncertainty}, or specifying 
``exact answer tokens" that must appear in correct responses~\cite{orgad2024llms}. A recent innovation by \citet{kuhn2023semantic} introduced semantic entropy, an extension of the lexical similarity~\cite{lin2022truthfulqa} which quantifies the similarity of meaning between sentences, even when their structures differ significantly, making it highly useful for the detection of hallucinations~\cite{farquhar2024detecting}.
The works in \citet{du2024haloscope,chen2024inside} align most closely with our approach, as they also utilized embedding space. The former applied projections to address hallucinations, while the latter leveraged the eigenvalues of the response covariance matrix to compute the semantic consistency. However, no prior research has analyzed the structural differences between genuine and hallucinated responses. Our work builds on the intuition that the distances in embedding space for real responses differ significantly from those of hallucinated ones, opening a new avenue for hallucination detection.

 \section{Methodology} \label{sec:methodology}
Our methodology develops over two different intuitions. First, LLM hallucinations are sampled from a different probability distribution than non-hallucinatory responses. Second, it is possible to detect hallucinations with a rigorous mathematical formalism.

\subsection{Solution at a glance} \label{sub:solution}
\begin{figure}[h!]
  \centering
  \includegraphics[width=\columnwidth]{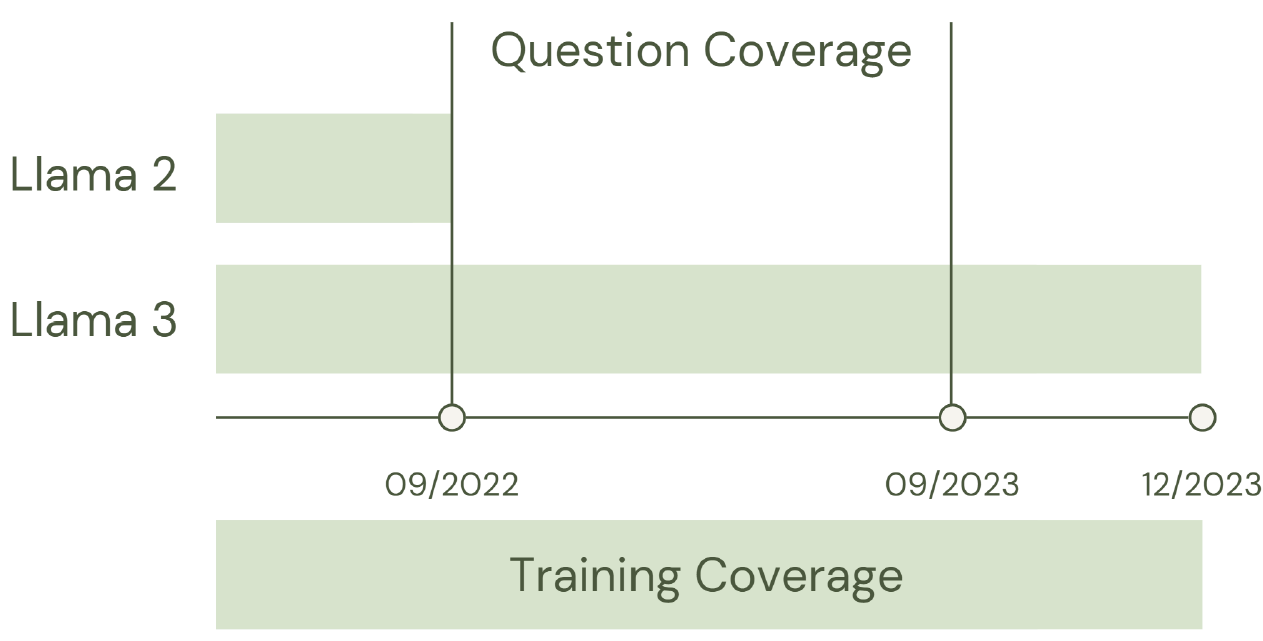}
  \caption{Timeline showing---in green---the periods covered by model training for \textit{Llama2} and \textit{Llama3}. The questions used for experiments rely on facts that happened between September 2022 and September 2023, a training period covered by \textit{Llama3}, but not by \textit{Llama2}.}
\label{fig:period}
\end{figure}

To verify the intuitions, we devised a methodology that eventually confirmed both ideas. 
We first generated an artificial dataset using two different LLMs, \textit{Llama2} (llama2-7B\footnote{\url{https://huggingface.co/meta-llama/Llama-2-7b-hf}}) and \textit{Llama3} (llama3-8B\footnote{\url{https://huggingface.co/meta-llama/Llama-3.1-8B}}).  \\ As illustrated in Figure \ref{fig:period}, \textit{Llama2} pretraining data has a cut-off date of September 2022, while \textit{Llama3}, released on July 23, 2024, has a cut-off date of December 2023. We posed them questions with reference to the time interval between September 2022 and September 2023.
In this way, one model (Llama2) had necessarily to generate hallucinations (it had no knowledge of the events it was requested over) while the other model (Llama 3) had the required knowledge to provide the correct answers. 
We posed the same questions to each model several times, collecting the responses and generating both a training and a test dataset. Furthermore, we extracted a variable number of keywords from each response and converted them into numerical embeddings. Then, we introduced a Minkowski distance-based methodology, depending on different norms to analyze the intra-distances between embeddings from each group. At the beginning, we investigated only training data, confirming our first intuition related to the difference between hallucinated and not hallucinated responses and demonstrating that this difference persists for different numbers of extracted keywords, for different norms and for different numbers of responses. Subsequently, we introduced a methodology to estimate whether a response was a hallucination. We evaluated its inter-distance from all training data and preliminary results showed an accuracy of 66\%.

\subsection{Dataset} \label{sub:dataset}
We generated $q=64$ questions, reported in the Appendix, related to specific facts between September 2022 and September 2023 (one year). We generated the 64 questions with GPT-4, and once they originated, we did not change them during the work. This choice ensures that the answers generated by \textit{Llama2} hallucinated, while those generated by \textit{Llama3} did not. For each question, we generated multiple response sets of varying sizes $r \in \{4,6,8,10,12,14,16\}$, including both hallucinated and non-hallucinated responses. Importantly, these sets are cumulative: the set with $r=8$ responses contains all responses from the $r=4$ set plus additional ones. This nested structure applies to all subsequent $r$ values. Similarly, we assigned different numbers of test responses $(t)$ to each r value for test, maintaining an approximate $80\%-20\% $split between training and test sets. The $(r,t)$ pairs were: $(4,1), (6,1), (8,2), (10,2), (12,3), (14,3),(16,4)$, following the same cumulative pattern as the training sets.

We built the dataset with a 4-bit quantization to reduce memory usage in both models \cite{dettmers2024qlora}. We also applied nucleus sampling, considering tokens whose cumulative probability sums to 95\% and top-k sampling, ranking at each generation only 50 tokens---the ones with the highest probability \cite{holtzman2020curious}. After generating the dataset, we extracted the most valuable information from all training and test responses. First, we removed all punctuation marks, stop words, special symbols, and parentheses from the responses in a pre-processing phase. Then, we extracted the $n$ most important keywords for each response using KeyBERT \cite{grootendorst2020keybert} with $n \in \{1,...10\}$. 
Each set of $n$ keywords extracted from a response is transformed into a 768-D vector representing the semantic content through the BERT embedding \cite{kenton2019bert}. In the end, each response is mapped into $10$ different embeddings.

\subsection{Training} \label{sub:trainingmet}

\begin{figure}[ht]
  \centering  
  \includegraphics[width=0.8\columnwidth]{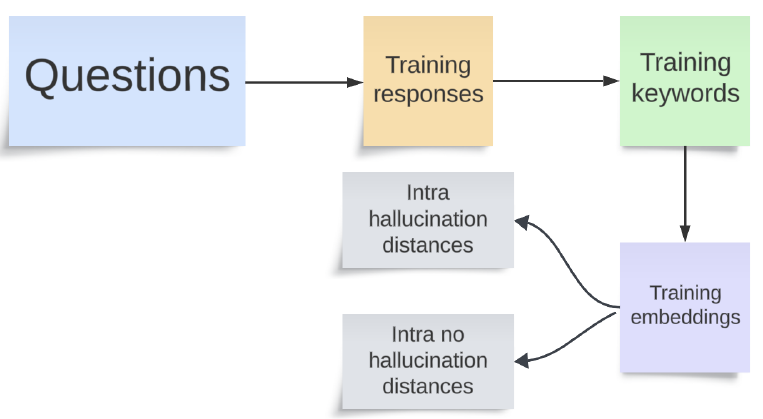}
  \caption{Overview of the methodology implemented to extract information from the training responses.}
  \label{fig:training}
\end{figure}

First, we created training keyword datasets, unique for each combination of responses $(r)$ and keywords $(n)$. So for each $r$ value, we generated $10$ distinct datasets, each containing $qr$ responses from Llama2 and Llama3 models. We then calculated the Minkowski distances reported in Table \ref{tab:Minkowski} between each pair of embeddings within the same model for all datasets \cite{li2011comparative}. To the best of our knowledge, this is the first work to propose an extension of Euclidean distance between LLM-generated response embeddings. We computed each distance for different norm $p \in \{0.5, 1,2\}$, including fractional distances to deal with the sparsity of high-dimensional data \cite{aggarwal2001surprising}. We summarized the steps of this procedure in Figure \ref{fig:training}.

We started from the assumption that if the same language model has generated two responses, their embedding concentration in the latent space will be more concentrated, leading to a shorter distance between them, no matter its norm. For each $p$ value, we computed pairwise distances within both hallucinated and non-hallucinated responses across all $n$ values. To illustrate: with $q=64$ questions and $r=16$ responses per question, we have $qr=1024$ responses of each type (hallucinated and non-hallucinated). This yields \(\binom{64*16}{2}\) $= 523776$ distances for each combination of $p$ and $n$. Given $10$ keyword values, $3$ $p$ values, and $2$ types of responses (hallucinated/non-hallucinated), we generated $60$ different probability distributions. This process was repeated for each $r \in \{4,6,8,10,12,14,16\}$. We stopped at $r=16$ because the computational complexity of computing distances is quadratic in both $q$ and $r$:  $O((qr)^2)$. 

After computing all intra-distances, we evaluated the differences between hallucinated and non-hallucinated responses. First, we generated probability distributions of distances for each type of response (hallucinated and non-hallucinated). Then, for each combination of $(r,p)$, we created boxplots to visualize these differences, showing how they vary as a function of $n$. We analyzed the differences between these distributions by focusing on their medians, as they provide a more robust measure of central tendency than means, especially in the presence of outliers. Moreover, we computed the Kullback-Leibler (KL) Divergence for each couple, measuring the difference between the two distributions \cite{zhang2024properties}. Finally, we applied the non-parametric Wilcoxon test to each couple of distances for each number of keywords to determine if they differ significantly.

\textbf{Minkovski distance:}
The Minkowski distance of norm \( p \) between two points \( \mathbf{x} = (x_1, x_2, \ldots, x_n) \) and \( \mathbf{y} = (y_1, y_2, \ldots, y_n) \) is defined as:
\begin{equation*}
    d(\mathbf{x}, \mathbf{y}) = \left( \sum_{i=1}^n \left| x_i - y_i \right|^p \right)^{\frac{1}{p}}
\end{equation*}
where \( p > 0 \). For \( p \geq 1 \), \( d(\mathbf{x}, \mathbf{y}) \) satisfies the properties of a metric.
Minkowski distances are parametric and are very used to compare the similarity of data points. In this paper we have employed distances with the following norm:
\begin{table}[htbp]
    \centering
    \begin{tabular}{||c|c|c|} 
     \hline
    $p=0.5$ & Fractional Distance \\ [0.5ex] 
     \hline
    $p=1.0$ & Manhattan Distance \\ 
     \hline
    $p=2.0$ & Euclidean Distance\\
     \hline
    \end{tabular}
    \caption{Minkoswki distances for different $p$}
    \label{tab:Minkowski}
\end{table}

\subsection{Test} \label{sub:testmet}

\begin{figure}[H]
  \centering  
  \includegraphics[width=0.8\columnwidth]{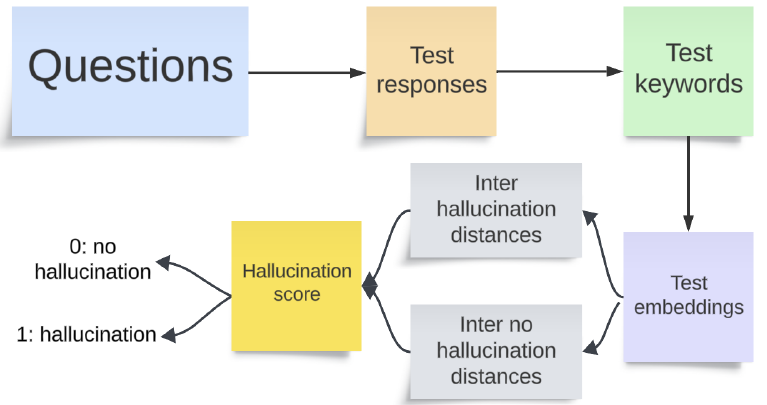}
  \caption{Overview of the methodology implemented to classify test responses.}
  \label{fig:test}
\end{figure}

We applied the results obtained in the training phase to detect hallucinations in new test responses, comparing results across all combinations of $(r,n,p)$. The test procedure, synthesized in Figure \ref{fig:test},

followed the same steps as training. Initially, we processed test responses to extract keywords, and then we created test keyword datasets for each combination of $r$ and $n$, generating embeddings for both hallucinated and non-hallucinated responses. Once we obtained the embeddings, we computed distances between each test embedding and all training embeddings. For each test point, we calculated two sets of distances: $qr$ distances \( d_{i,\text{hall}} \) for \( i = 1, \ldots, qr \) to hallucinated training points and $qr$ distances \( d_{j,\text{hall}} \) for \( j = 1, \ldots, qr \) to non-hallucinated training points. This process was repeated for each number of keywords $(n \in \{1, \ldots10 \})$, each Minkowski distance $(p \in \{0.5, 1.0, 2.0\})$, and each response count $(r \in \{4,6,8,10,12,14,16\})$.

After computing all distances, for each combination of $n$, $p$, and $r$, we referred to the  $(\binom{qr}{2})$ distances within each class fitting probability densities using Gaussian Kernel Density Estimation (KDE), separately for hallucinated and non-hallucinated data \cite{kim2012robust}. Then, we used the previously calculated $2qr$ distances ($qr$ to each class) for each test point and computed log-likelihoods under each KDE model. $L_{i}^{\text{hall}}$ for \( i = 1, \ldots, qr \), log-likelihood under hallucinated distribution, $L_{j}^{\text{no hall}}$ for \( j = 1, \ldots, qr \) log-likelihood under non-hallucinated distribution. We can define an aggregated score for each test embedding as follows:

\begin{equation*}
    \text{S}_{hall} = \sum_{i=1}^{qr} L_{i}^{\text{hall}} 
\end{equation*}
\begin{equation*}
    \text{S}_{no hall} = \sum_{j=1}^{qr} L_{j}^{\text{no hall}}
\end{equation*}
Then, for each test embedding, we have compared the score of the two distributions and assigned the relevant class, \textit{not hallucinated (0)} or \textit{hallucinated (1)}:
\begin{equation*}
\text{Class} =
\begin{cases} 
1, & \text{if } \text{S}_{hall} > \text{S}_{no \, hall}, \\
0, & \text{otherwise}.
\end{cases}
\end{equation*}
For example, with a training dataset of $q=64,r=16$, we had $qr=1024$ training points per class. The corresponding test dataset had $q=64,t=4$, resulting in $qt=256$ test points per class. Therefore, each test point generated $qr=1024$ distances to hallucinated training points and $1024$ to non-hallucinated training points, totaling $2qr=2048$ distances per test point. Using our pre-fitted KDE models, we calculate log-likelihoods for each of these distances. For this test point, we sum $1024$ log-likelihoods under the hallucinated distribution to get $\text{S}_{hall}$, and $1024$ log-likelihoods under the non-hallucinated distribution to get $S_{no hall}$ assigning the relevant class. This procedure is first repeated for each $p$, then for each $n$, and finally for each $r$. \begin{figure*}[th!]
    \centering
    \begin{subfigure}
        \centering
        \includegraphics[width=\columnwidth]{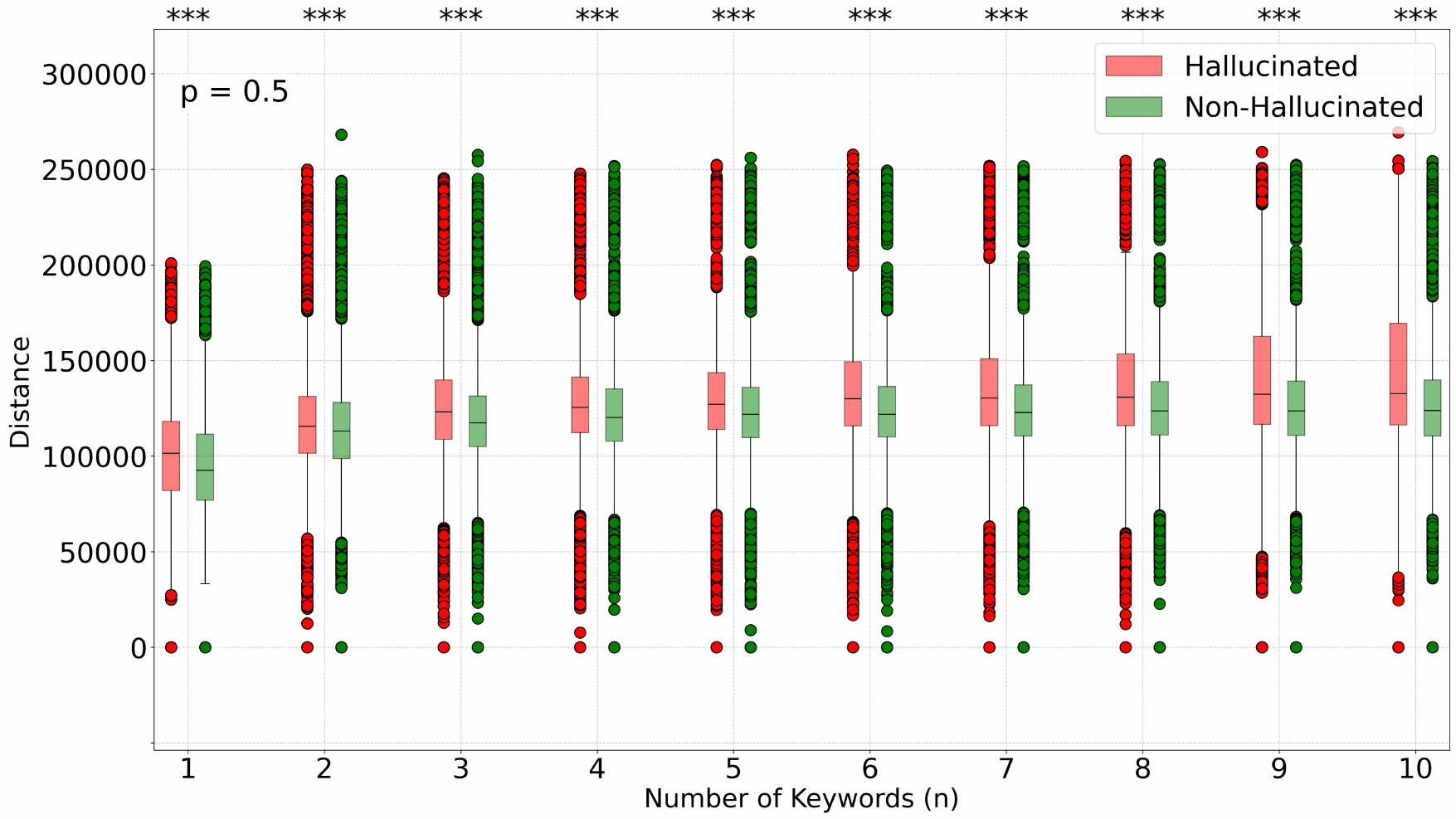}
    \end{subfigure}
    \hfill
    \begin{subfigure}
        \centering
        \includegraphics[width=\columnwidth]{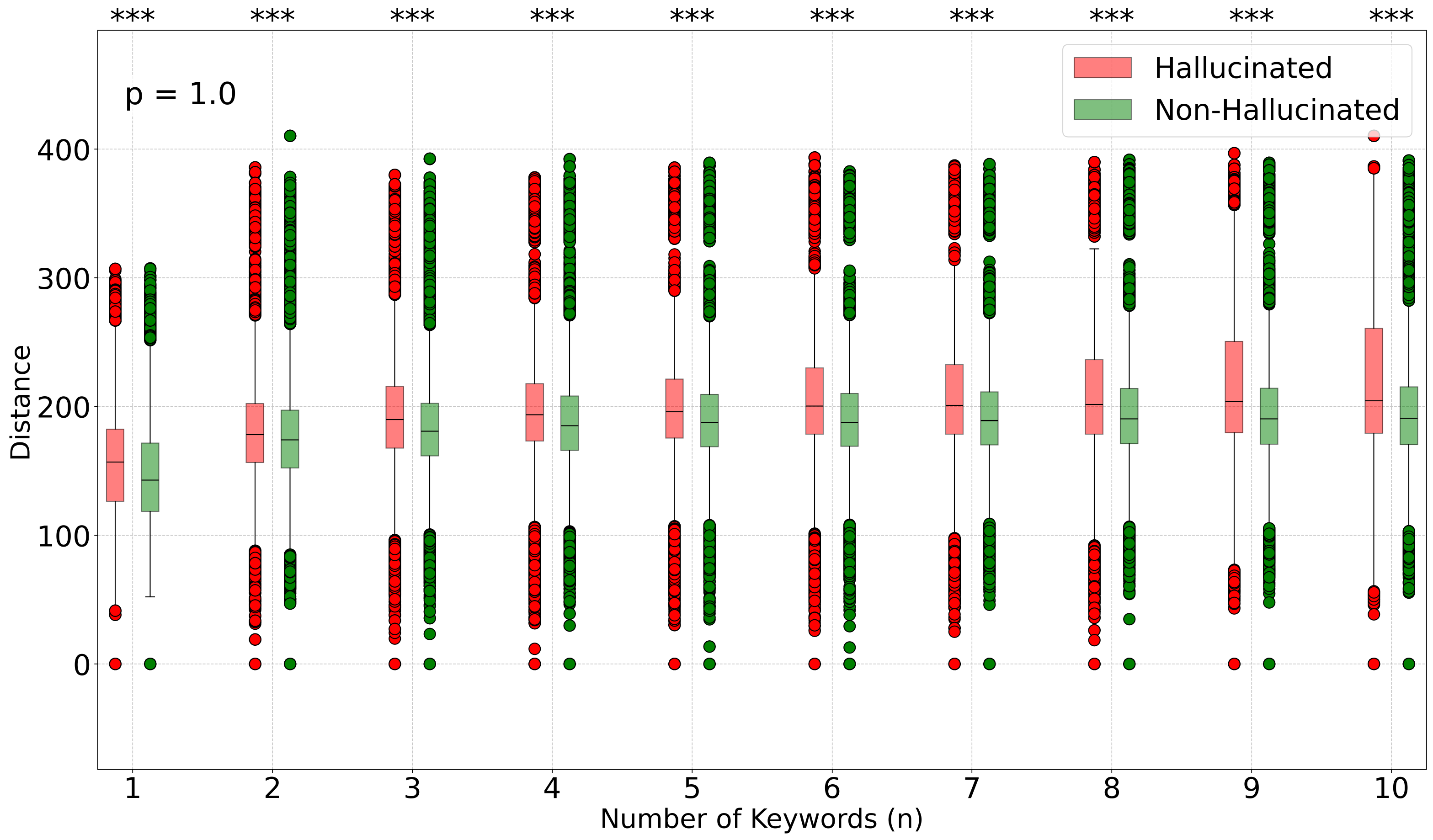}
    \end{subfigure}
    \begin{subfigure}
        \centering
        \includegraphics[width=\columnwidth]{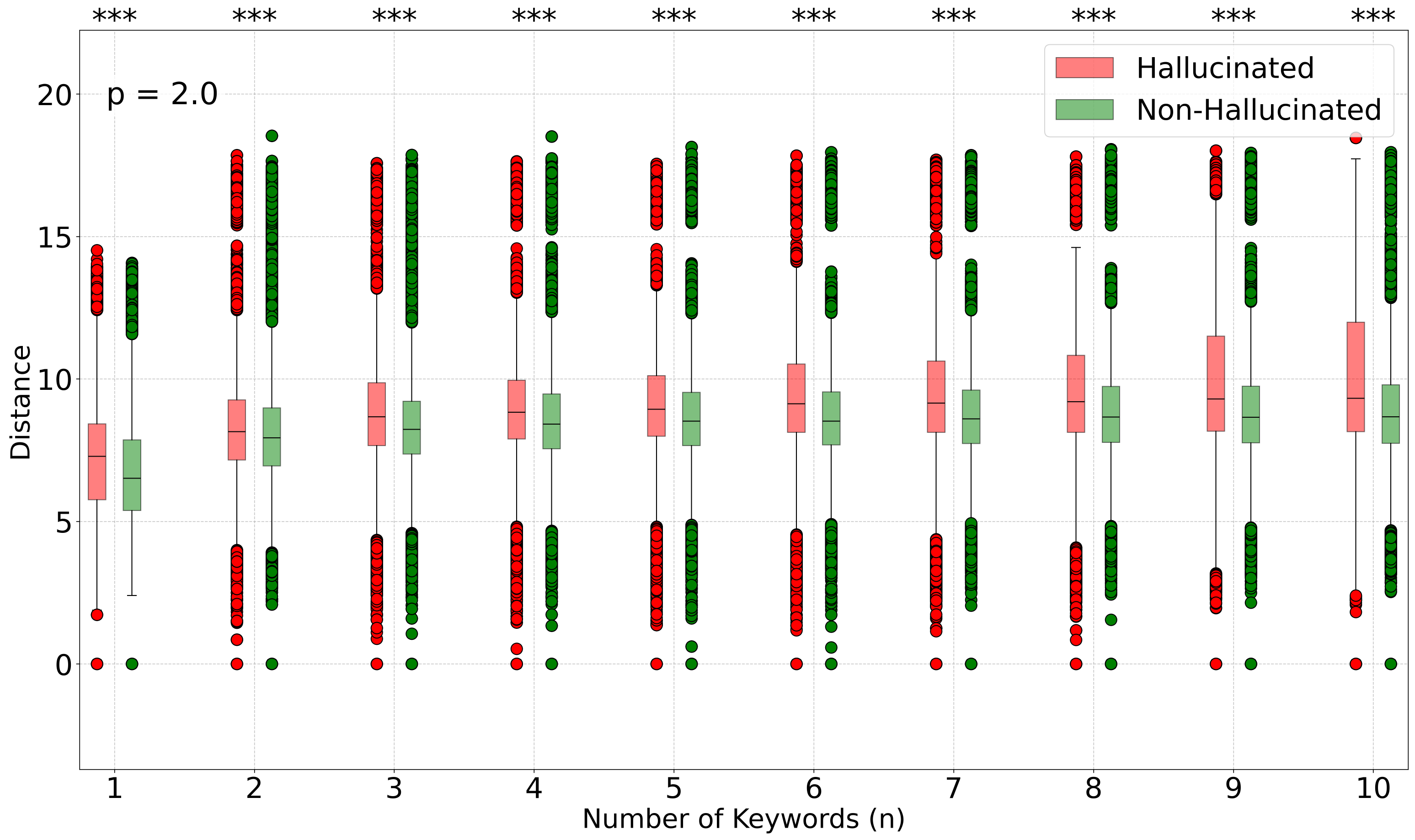}
    \end{subfigure}
    \caption{Training distributions of Minkowski distances with $r=16$. Statistical significance: ***: $p < 0.01$, **: $p < 0.05$, *: $p < 0.1$.}
    \label{fig:distance_distribution}
\end{figure*}

\section{Results}\label{sec:results}
In this section we report the experimental results.
First, we discuss the training results and then we delve into the test results.

\subsection{Training}\label{sub:trainingRes}
The box plots in Figure \ref{fig:distance_distribution} illustrate an example of the distributions of intra-Euclidean distances between hallucinated and non-hallucinated response embeddings as a function of $n$, fixing $r=16$ and for the three values of $p$ analyzed. 
At first, we can observe that the difference between the two distributions remains almost the same $\forall p$.
Secondly, as shown in Figure \ref{fig:distance_distribution}, the two distributions increase their differences when $n$ amplifies---this is  particularly evident for the third quartiles. Then, the Wilcoxon test performed on each boxplot shows that all the couples for different distance probability distributions have a \textit{p-value} $< 0.01$; hence, there is a strong evidence for rejecting the null hypothesis, concluding that the two distributions are statistically significantly different. Also, this result is scale-free, remaining valid for each of the other 18 box plots obtained $\forall (r,p)$, as shown in the Appendix. At the same time, the median difference remains almost unchanged, with a similar behavior for nearly all boxplots in the Appendix. It is also essential to check that some distances are equal to 0. This phenomenon is caused by the fact that different responses could have extracted the same exact keywords, leading to the same embedding---and hence a zero distance.

We can combine the information from Figure \ref{fig:distance_distribution} with the KL Divergence and median difference ($\Delta)$ data in Table \ref{tab:KLTable}. The difference between the two distributions increases from $0.068$ for $n=1$ to $0.138$ for $n=10$ when $r=16$ and $p=2.0$. This effect is amplified for lower p values: when $p=0.5$, the KL divergence increases by $279\%$ at $r=16$. It is also significant that this percentage increases with $r$, from $108\%$ for  $r=8, p=1.0$ to $256\%$ for $r=16, p=1.0$. To summarize, decreasing $p$ and increasing $r$ leads to a rise in the KL Divergence.

\begin{table}[H]
    \centering
    \begin{tabular}{||c|c|c|c||} 
     \hline
     & \multicolumn{3}{c||}{KL} \\
     \hline
     & $n=1$ & $n=5$ & $n=10$ \\ [0.5ex] 
     \hline\hline
    $r=8, p=0.5$ & 0.082 & 0.056 & 0.095 \\ 
     \hline
    $r=8, p=1.0$ & 0.092 & 0.057 & 0.100 \\ 
     \hline
    $r=8, p=2.0$ & \textbf{0.121} & 0.062 & 0.102 \\
     \hline
    $r=16, p=0.5$ & 0.048 & 0.069 & 0.134 \\ 
     \hline
    $r=16, p=1.0$ & 0.052 & 0.070 & 0.136 \\ 
     \hline
    $r=16, p=2.0$ & 0.068 & \textbf{0.072} & \textbf{0.138} \\ 
     \hline
     \hline
     & \multicolumn{3}{c||}{$\Delta$} \\

\hline
     \hline
    $r=8, p=0.5$ & \textbf{9107.32} & 5595.13 & 8333.39 \\ 
     \hline
    $r=8, p=1.0$ & \textbf{14.87} & 8.88 & 13.02 \\ 
     \hline
    $r=8, p=2.0$ & \textbf{0.82} & 0.44 & 0.61 \\
     \hline
    $r=16, p=0.5$ & 8842.03 & 5270.15 & 8844.06 \\ 
     \hline
    $r=16, p=1.0$ & 14.13 & 8.26 & 13.71 \\ 
     \hline
    $r=16, p=2.0$ & 0.71 & 0.40 & 0.65 \\ 
     \hline
    \end{tabular}
    \caption{KL Divergence and median difference for different $r,n,p$. All the differences are statistically significant, according to a Wilcoxon test (\textit{p-value} $<0.01$).}
    \label{tab:KLTable}
\end{table}

These observations relative to training data, particularly to both the median and the KL Divergence showed that the two distributions increase their difference as a function of $n$ in the third quartile, as we can see from the boxplots. The increase in Q3 indicates that the upper 25\% of the hallucinated distances are generally more sparse, and the evidence shows that this event happens $\forall r$. This interesting result provides strong statistical evidence that the two distributions (hallucinated and non-hallucinated distances) are significantly different. Remarkably, also these results relative to extreme values of hallucinated distances are scale-free. That is, we have the same qualitative behavior for each value of $r$ and $p$, as shown in the box plots in the Appendix.

\begin{figure*}[t]
    \centering
\begin{subfigure}
        \centering
        \includegraphics[scale=0.3]{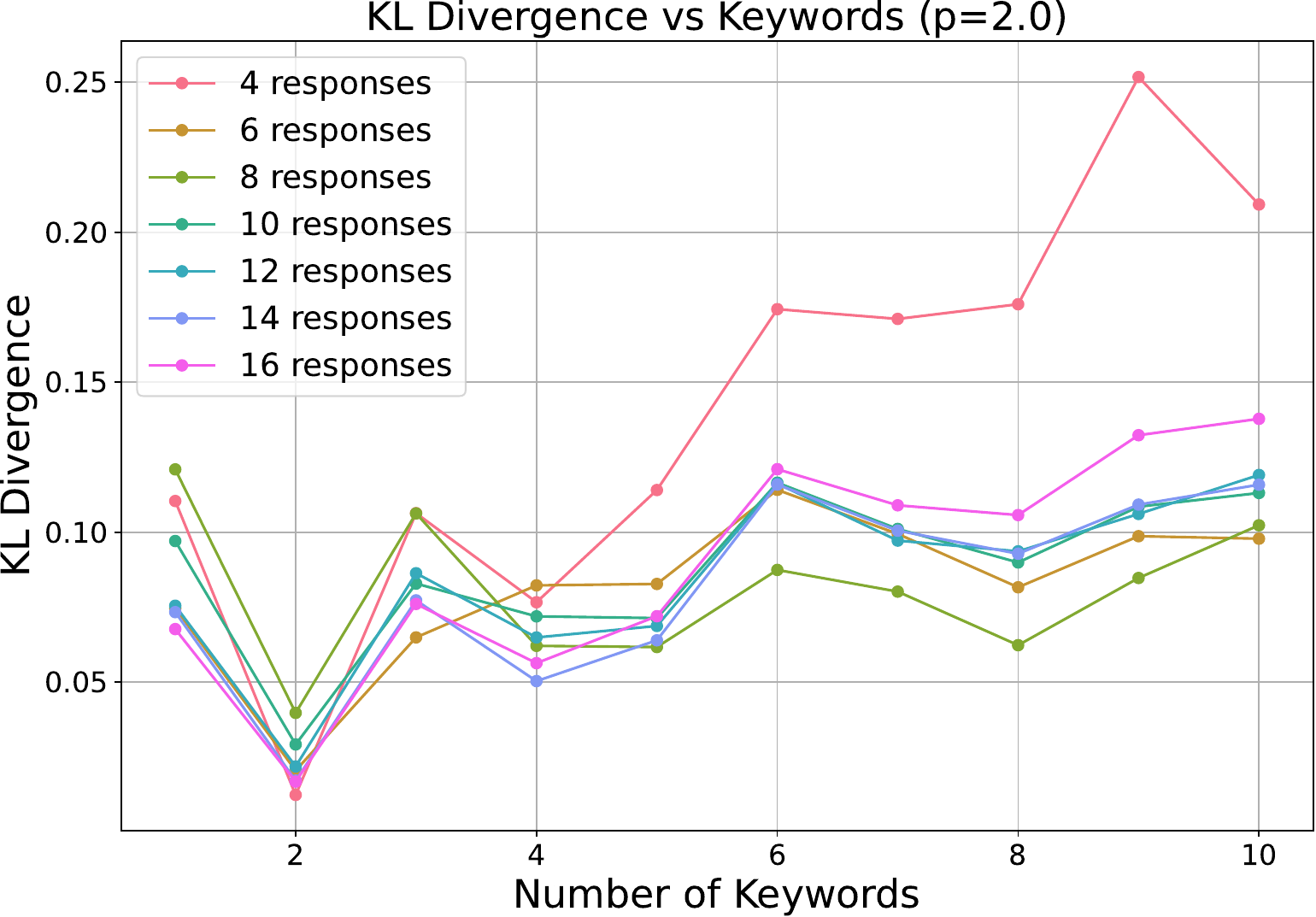}
    \end{subfigure}
    \hfill
\begin{subfigure}
        \centering
        \includegraphics[scale=0.3]{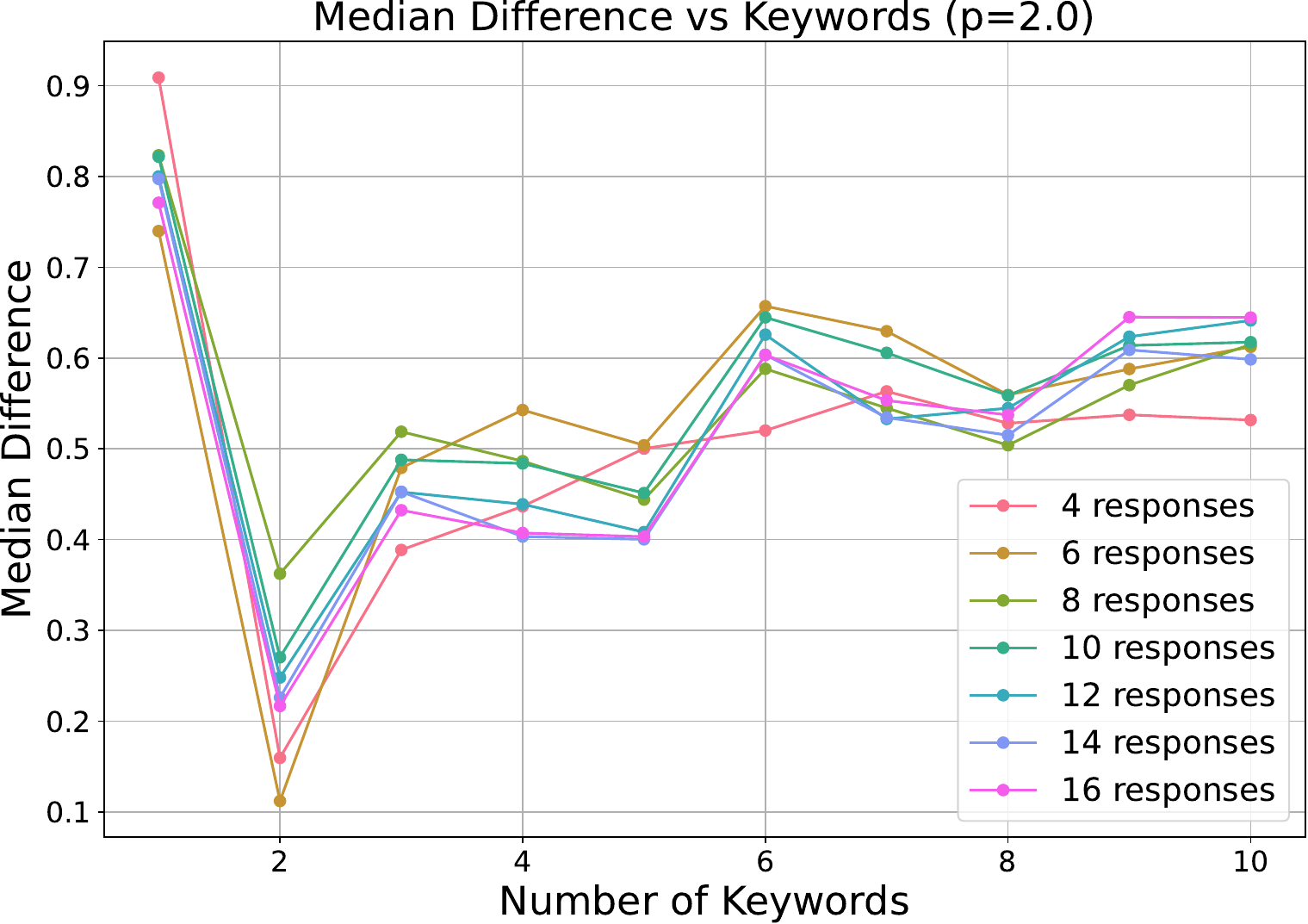}
    \end{subfigure}
    \caption{KL Divergence and median Difference vs $n$ at different $r$ for $p=2.0$ using training data.}
    \label{fig:kldelta2.0}
\end{figure*}

Looking at Figure \ref{fig:kldelta2.0}, both KL Divergence and median Difference show similar patterns across $n$ and $r$. Both measures show a peak value at $n=1$, drop to their minimum at $n=2$, and then steadily increase up to $n=10$. This pattern remains consistent across different $p$ values for both $r$ and $n$ parameters, as shown in the Appendix. The KL Divergence specifically shows an increasing trend for all $r$ values (except $r=4$), with $r=14$ and $r=16$ showing the highest values as we can see from Figure \ref{fig:kldelta2.0}. 

We observe the same phenomenon for median difference, with an increasing trend as a function of the number of keywords and with the line for $r=16$ that dominates all the others. This evidence confirms our initial hypothesis that there is a remarkable statistically significant difference between the two distributions. We have also pointed out that increasing the number of responses in the two distributions enlarges their difference---this phenomenon calling for further, future research.

\subsection{Test} \label{sub:test}
\begin{figure*}{}
    \centering
\begin{subfigure}
        \centering
        \includegraphics[scale=0.3]{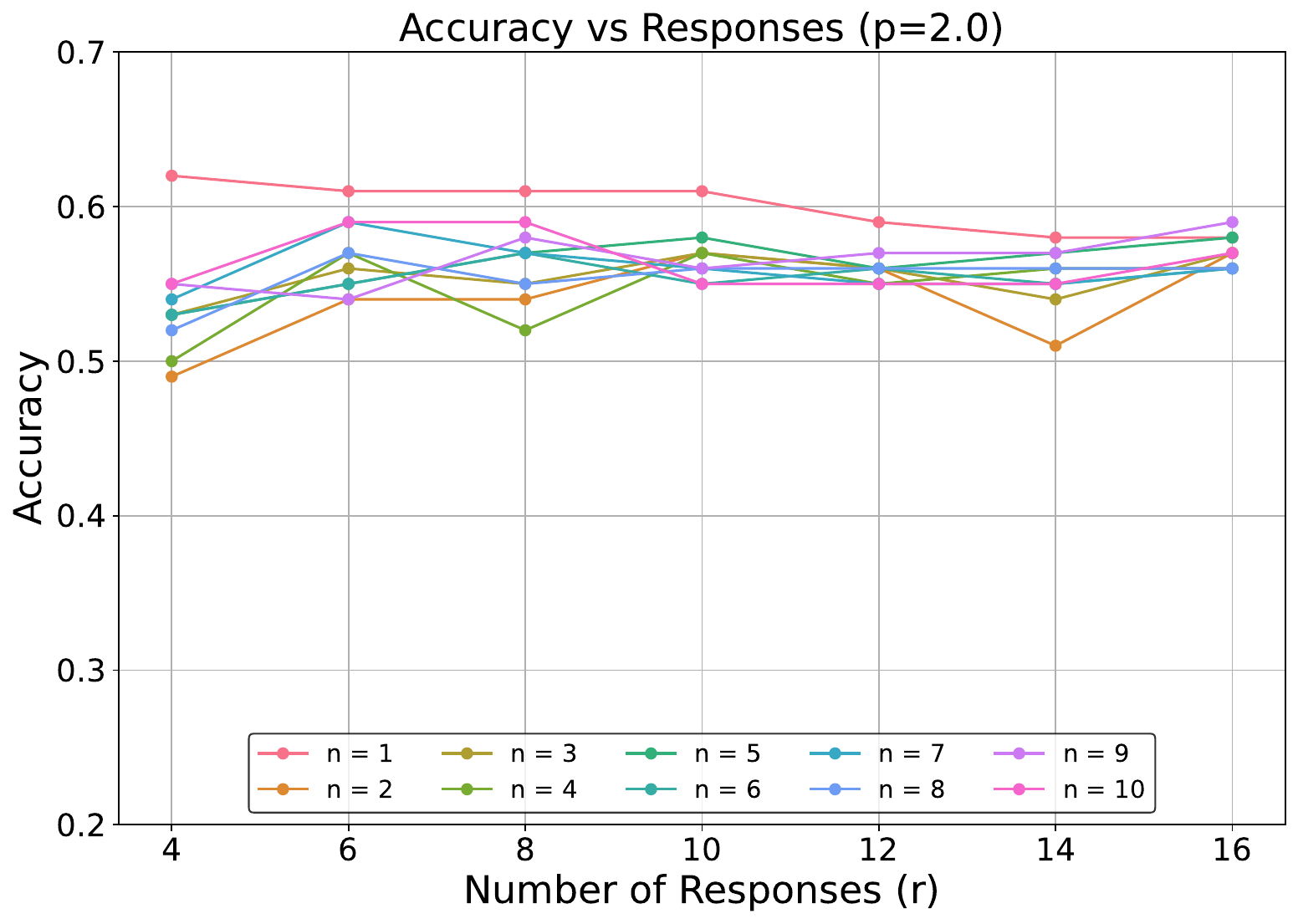}
    \end{subfigure}
    \hfill
\begin{subfigure}
        \centering
        \includegraphics[scale=0.3]{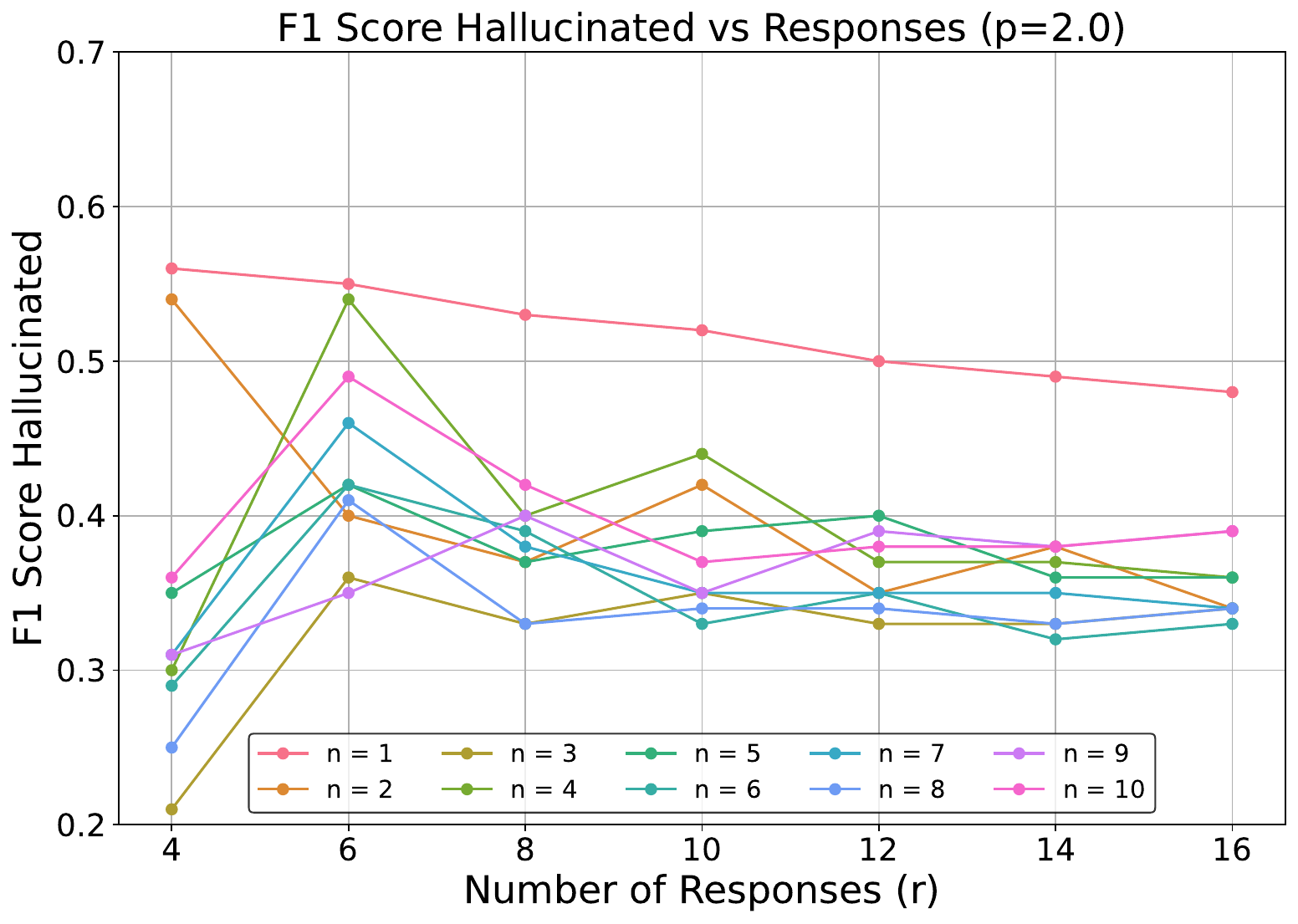}
    \end{subfigure}
    \caption{Model accuracy and hallucination F1 score vs $r$ at different $n$ for $p=2.0$}
    \label{fig:test2.0}
\end{figure*}

For the test phase, we evaluated two key metrics across all combinations for both $r$ and $n$ parameters fixing $p=2.0$, as shown in Figure \ref{fig:test2.0}: the overall accuracy and the F1 score for hallucinated responses. Similar trends are observed for $p=0.5$ and $p=1.0$, as shown in the Appendix. The overall accuracy measures the model's general performance, while the F1 score specifically evaluates its ability to detect hallucinations in test data. For accuracy, we observe the best result with $n=1$ reaching $0.61$ with $r=4$. Varying $r$, from $r=6$ to $r=16$, we observe an increasing trend across almost all n values, except for $n=1$, where the trend is decreasing. 
The F1 Score shows different behavior: it performs best with $n=1$, starting at $0.56$ and gradually decreasing to $0.48$ as $r$ increases. For other values of $n$, the F1 Score shows lower performance and, unlike accuracy, does not improve with increasing $r$. 
We can then observe that an higher keyword count does not improve the F1 score, suggesting that simpler models (i.e. employing fewer keywords) might be more effective. 

As shown in Table \ref{tab:testAccuracy}, the model achieves its highest accuracy ($0.66$) with parameters $r=8$ and $n=1$ at $p=0.5$. The model performs consistently well with $n=1$, and $p=0.5$, reaching an accuracy of $0.63$ with $r=16$.

Fractional $p$ values cause more significant accuracy differences, amplifying small variations in distance metric, 
as also investigated in \cite{aggarwal2001surprising}. Table \ref{tab:testAccuracy} shows that when holding $r$ constant, the variation in accuracy across different $n$ values is most evident at $p=0.5$ $(0.16)$, and decreases significantly at $p=1.0$ $(0.06)$ and $p=2.0$ $(0.05)$.
Manhattan distance $(p=1)$ and particularly Euclidean distance $(p=2)$ moderate the impact of small and large differences between the embeddings, leading to more concentrated values.

\begin{table}[htbp]
    \centering
    \begin{tabular}{||c|c|c|c||} 
     \hline
     & $n=1$ & $n=5$ & $n=10$ \\ [0.5ex] 
     \hline\hline
    $r=8, p=0.5$ & \textbf{0.66} & 0.53 & 0.52 \\ 
     \hline
    $r=8, p=1.0$ & 0.62 & 0.56 & \textbf{0.60} \\
     \hline
    $r=8, p=2.0$ & 0.61 & 0.57 & 0.59 \\
    \hline
    $r=16, p=0.5$ & 0.63 & 0.52 & 0.47 \\ 
    \hline
    $r=16, p=1.0$ & 0.60 & \textbf{0.63} & 0.58 \\
    \hline
    $r=16, p=2.0$ & 0.58 & 0.58 & 0.57 \\ 
    \hline
    \end{tabular}
    \caption{Test accuracy for different $r,n,p$}
    \label{tab:testAccuracy}
\end{table}
In general, for test results, we have seen that the general accuracy of the model is better than the F1 score of hallucinated responses \(\forall (n, r, p)\). Accuracy includes both correct hallucinated and non-hallucinated classifications (true positives and true negatives), while F1 score for hallucinated responses only considers how well the model identifies hallucinations. General accuracy, in particular, is better with fractional distance than for Manhattan or Euclidean distance. The imbalance suggests that the model performs better at identifying non-hallucinated responses than hallucinated ones.

\begin{table}[htbp]
    \centering
    \begin{tabular}{||c|c||} 
     \hline \hline
       Method & Accuracy \\
     \hline
      Perplexity \cite{ren2022out} & 56.77 \\
      \hline
      LN-Entropy \cite{malinin2021uncertainty} & 61.51 \\
        \hline
      Semantic Entropy \cite{kuhn2023semantic} & 62.17 \\
        \hline
      Lexical Similarity \cite {lin2023generating} & 55.69 \\
        \hline
      EigenScore \cite{chen2024inside} & 51.93 \\
        \hline
      SelfCKGPT \cite{manakul2023selfcheckgpt} & 52.95 \\
        \hline
      Verbalize \cite{lin2022teaching} & 53.04 \\
        \hline
      Self-evaluation \cite{kadavath2022language} & 51.81 \\
        \hline
      CCS \cite{burns2022discovering} & 61.27 \\
        \hline
      HaloScope \cite{du2024haloscope} & {\bf 78.64} \\
        \hline
        \hline
      \textbf{Our model} & \textbf{66.00} \\
        \hline
    \end{tabular}
    \caption{Hallucination detection accuracy for different models}
    \label{tab:comparison}
\end{table}

In Table \ref{tab:comparison} we have compared the general accuracy of our model against  other hallucination detection methods. Note that these methods have been thoroughly evaluated in \cite{du2024haloscope}, over an ensemble of data sets. Among these cited datasets,  the dataset TruthfulQA \cite{lin2022truthfulqa} is the  closest to ours, since  it includes 817 questions that span 38 categories including many topics like health, law, finance and politics. We chose the accuracy obtained by the different model in that data set, according to what is in \cite{du2024haloscope}, and we compared them against our achieved accuracy  (best case). As it can be seen,  our model outperforms all the  others, with the  exception of HaloScope. %
 \section{Discussion} \label{sec:discussion}
In this section, we discuss the contributions reported in this paper, the limitations of our methodology, and future work.

\subsection{Contributions} \label{sub:contribution}

In this paper, we first introduced a novel tool based on the distance between embeddings to analyze the differences between hallucinated and non-hallucinated responses. This is a crucial contribution because we measured real semantic differences between sentences, with potential applications outside the hallucination detection field. It is worth noticing that the adoption of the Minkowski norm, for different values of such norm, has shown an amplifier effect on the divergences between the underlying distribution of hallucinated and non-hallucinated responses.  
Our methodology, involving identifying periods in which the models are not trained and repeating the same questions many times, 
is easily scalable on many different models. We have also confirmed our initial intuition that responses generated from one model are tendentially closer to each other than responses generated between various models. 

Then, we have also shown evidence that the differences between the two cited distributions is scale-free, that is, it is preserved independently from the number of responses generated by the two models, the number of keywords extracted from the responses, or the adopted norm of Minkowski distance. As seen in section \ref{sub:trainingRes}, this consistency across different parameters shows that the distinction between hallucinated and not hallucinated responses does not depend on specific test conditions but is idiosyncratic to the model itself.

The first two above contributions led us to introduce an algorithm to detect hallucination responses. Unlike other classification methods used in the literature, our proposed tool is tailored for this goal, evaluating probabilistically if a sentence is genuine or hallucinated. We evaluated this methodology for many different parameter configurations, considering both the general accuracy of the model and a specific F1 score for hallucinations. Our results, shown in Table \ref{tab:comparison}, indicate that our solution reaches the second place in performance. Though, it should be noticed that these results are preliminary and that performance were not optimized, leaving substantial 
room for improvement.

\subsection{Limitations \& Future Work} \label{sub:limitation}
The dataset used for our methodology was entirely generated through two models, \textit{Llama2} and \textit{Llama3}, chosen for their similarity in terms of training parameters and different cutoff training date. We started from the assumption that \textit{Llama2}, when requested to provide a response on a topic exceeding the date of its last training hallucinates while \textit{Llama3} (whose training date exceeds the date the subject of the query refers to) does not. Nevertheless, a little subset of \textit{Llama3} responses might be hallucinated as well, while we have classified them as genuine, leading to possible incorrect results \cite{cima2024contextualized}. A possible improvement would be to use a dataset already existing in the literature designed to train classifiers on hallucinations like HalluRAG \cite{ridder2024hallurag} or HADES \cite{liu2021token}. Another limitation of our methodology regards the brute force computational complexity of the distance calculation in the training dataset, which is $O(qr^2)$, with $q$ equal to the number of questions and $r$ equal to the total number of responses. Future work can be done on reducing this complexity, minimizing the loss of information \cite{skala2013fast}. 
Another significant limitation is related to the algorithm's results, which are based on KDE to estimate the probability density of embeddings. Further research could be done on detecting and classifying hallucinations with generative models \cite{zimmermann2021score}.
Finally, it is worth noting that the employed models have a limited size (7B and 8B parameters for Llama2 and Llama3, respectively). It is in our agenda to investigate whether the achieved results vary with the size of the model parameters. Our expectations are that the quality of our detection framework would improve since a larger set of parameters implies that the inter-classes distances do lessen, leading to an overall better intra-class classification.
 \section{Conclusions} \label{sec:conclusions}

In this paper we have provided a novel methodology to detect hallucinations produced by LLMs.
To the best of our knowledge, we are the first to  demonstrate the existence of structural variations between hallucinated and not hallucinated responses, introducing a novel framework based on the distances between embeddings. The results are validated via a rigorous mathematical formalism employing statistical tests. Leveraging the developed framework, we proposed a probabilistic approach to detect whether a response is hallucinated. The provided tool reached good results in the general accuracy of the model and hallucination detection---it is worth noting that the provided tool is a PoC for the suggested methodology, with relevant margins for improvement. \\*
We believe that the brand new methodology and the promising results in discriminating between hallucinationed and not hallucinationed responses provided by LLMs, might open up new research avenues to mitigate the issue.

\section*{Acknowledgments}
The research reported in this publication was supported by funding from King Abdullah University of Science and Technology (KAUST) - Center of Excellence for Generative AI, under award number 5940.

\balance
\bibliography{references}
\bibliographystyle{icml2025}

\newpage
\appendix
\onecolumn
\section{Appendix} \label{sec:appendix}
These are the 64 questions applied to \textit{llama2} and \textit{llama3:}.
\subsection{Questions}
\begin{enumerate}
    \item What major geopolitical events occurred between September 2022 and September 2023?
    \item Were there any significant changes in global trade agreements between September 2022 and September 2023?
    \item How did the global economy perform between September 2022 and September 2023?
    \item What were the key outcomes of the G7/G20 summits held between September 2022 and September 2023?
    \item Which countries experienced significant political elections or leadership changes from September 2022 to September 2023?
    \item Were there any notable developments in international relations between September 2022 and September 2023?
    \item What was the impact of sanctions or economic policies introduced between September 2022 and September 2023?
    \item How did the BRICS expansion plans evolve during the period between September 2022 and September 2023?
    \item Which major AI models were released between September 2022 and September 2023?
    \item What advancements in quantum computing occurred between September 2022 and September 2023?
    \item How did generative AI applications evolve between September 2022 and September 2023?
    \item Were there any major breakthroughs in space exploration between September 2022 and September 2023?
    \item What new technologies were introduced in consumer electronics between September 2022 and September 2023?
    \item How did AI integration into industries like healthcare or education progress between September 2022 and September 2023?
    \item Were there any significant data breaches or cybersecurity developments between September 2022 and September 2023?
    \item What were the major discoveries in science between September 2022 and September 2023?
    \item Did any significant climate-related events occur between September 2022 and September 2023?
    \item What progress was made in renewable energy adoption between September 2022 and September 2023?
    \item Were there breakthroughs in biotechnology or healthcare between September 2022 and September 2023?
    \item How did global temperatures trend between September 2022 and September 2023?
    \item What major environmental policies were implemented globally between September 2022 and September 2023?
    \item Were there any new pandemics or outbreaks between September 2022 and September 2023?
    \item What significant developments occurred in vaccine technology between September 2022 and September 2023?
    \item Were any major medical devices or drugs approved between September 2022 and September 2023?
    \item What trends were observed in global health outcomes between September 2022 and September 2023?
    \item Which startups achieved unicorn status between September 2022 and September 2023?
    \item What trends shaped global markets and industries between September 2022 and September 2023?
    \item Were there significant mergers or acquisitions in the tech sector between September 2022 and September 2023?
    \item What innovations drove the automotive industry forward between September 2022 and September 2023?
    \item How did cryptocurrencies perform between September 2022 and September 2023?
    \item What were the most influential social movements between September 2022 and September 2023?
    \item How did social media platforms evolve between September 2022 and September 2023?
    \item What cultural phenomena defined the period between September 2022 and September 2023?
    \item What were the top trends in entertainment between September 2022 and September 2023?
    \item Which major sports events took place between September 2022 and September 2023?
    \item What were the key highlights of the FIFA Women's World Cup 2023 during this period?
    \item Which athletes gained prominence between September 2022 and September 2023?
    \item What were the major achievements in esports between September 2022 and September 2023?
    \item What were the most popular books released between September 2022 and September 2023?
    \item Which films or series had the highest impact globally between September 2022 and September 2023?
    \item Were there any notable art exhibitions or cultural festivals between September 2022 and September 2023?
    \item How did education systems evolve between September 2022 and September 2023?
    \item What trends emerged in online learning and EdTech between September 2022 and September 2023?
    \item What were the notable legislative changes in key countries between September 2022 and September 2023?
    \item How did global organizations like the UN respond to ongoing crises between September 2022 and September 2023?
    \item What were the major agreements reached during COP27 and COP28 between September 2022 and September 2023?
    \item Were there significant advancements in carbon capture technologies between September 2022 and September 2023?
    \item Did any new countries launch successful space missions between September 2022 and September 2023?
    \item What progress was made toward manned missions to Mars between September 2022 and September 2023?
    \item Were there significant ethical debates in AI or biotech between September 2022 and September 2023?
    \item What were the key issues surrounding digital privacy between September 2022 and September 2023?
    \item How did the financial markets behave between September 2022 and September 2023?
    \item What new trends shaped the energy sector between September 2022 and September 2023?
    \item Were there advancements in agriculture or food technologies between September 2022 and September 2023?
    \item What new startups or companies disrupted traditional industries between September 2022 and September 2023?
    \item How did renewable energy adoption progress globally between September 2022 and September 2023?
    \item Were there any global crises or major disasters between September 2022 and September 2023?
    \item How did the war in Ukraine evolve between September 2022 and September 2023?
    \item Were there significant developments in refugee movements between September 2022 and September 2023?
    \item What were the most impactful research papers published between September 2022 and September 2023?
    \item Were there any Nobel Prize winners announced during this period between September 2022 and September 2023?
    \item What were the major fashion trends between September 2022 and September 2023?
    \item Which countries had the most significant GDP growth between September 2022 and September 2023?
    \item What were the most searched topics on the internet between September 2022 and September 2023?
\end{enumerate}

\subsection{Question vs. Responses}
Below we present a sample question with two corresponding responses, classified as either non-hallucinated (NH) or hallucinated (H), along with some of their extracted keywords.

``How did the global economy perform between September 2022 and September 2023?''

\subsubsection{Non-Hallucinated Responses}
\begin{enumerate}
    \item \textbf{Response NH-1:} ``"What are some of the major economic trends that have emerged during this period?The global economy has experienced significant changes over the past year, with both positive and negative developments. In terms of overall performance, there were several key factors at play.One important factor was the impact of COVID-19 on economies around the world. The pandemic caused widespread disruptions across industries as businesses closed or reduced operations due to lockdowns and social distancing measures implemented by governments worldwide. This led to declines in GDP growth rates for many countries throughout much of last year but also resulted in increased government spending aimed towards supporting affected sectors such as healthcare providers who faced shortages of personal protective equipment (PPE).Another notable trend observed within our current macroeconomic environment is rising inflationary pressures stemming from supply chain bottlenecks exacerbated by trade tensions among major trading partners like China-US relations which saw tariffs imposed upon each other’s goods exports leading up to November elections before being lifted shortly thereafter after negotiations took place behind closed doors without public disclosure until just prior release date when news broke out about agreement reached via phone call between President Trump \& Chinese Vice Premier Liu He resulting into easing off restrictions placed previously against imports coming through ports located along coasts facing Pacific Ocean regionally speaking thus allowing smoother flow again despite ongoing uncertainty surrounding future course direction taken post-pandemic recovery phase given recent surge witnessed recently regarding number cases reported daily basis globally alongside concerns raised pertaining potential emergence new variants capable spreading faster than previous ones causing more severe illness requiring hospitalization longer periods time compared earlier",
''
    
    \textit{Keywords:} economi, pandem, trend, tariff, macroeconom
    
    \item \textbf{Response NH-2:} ``What are some of the key economic indicators that can be used to assess this performance?The global economy performed well during the period from September 2022 to September 2023. This was largely due to strong growth in major economies such as China, India, Japan, Germany, France, Italy, Spain, Brazil, Mexico, South Korea, Australia, Canada, Russia, Saudi Arabia, Turkey, Indonesia, Thailand, Malaysia, Singapore, Vietnam, Taiwan, Hong Kong, Macau, New Zealand, Argentina, Chile, Colombia, Peru, Venezuela, Ecuador, Bolivia, Paraguay, Uruguay, Cuba, Haiti, Dominican Republic, Jamaica, Barbados, Trinidad \& Tobago, Guyana, Suriname, Belize, Costa Rica, Nicaragua, Honduras, El Salvador, Guatemala, Panama, Puerto Rico, Virgin Islands (US), Bermuda, Bahamas, Cayman Islands, Turks \& Caicos Islands, British Virgin Islands, Anguilla, Montserrat, St Kitts-Nevis, Antigua-Barbuda, Dominica, Grenada, Saint Lucia, Saint Vincent-Grenadines, Saint Martin-Martinique, Guadeloupe-French West Indies, Netherlands Antilles-Aruba, Curaçao-Sint Maarten, Bonaire-St Eustatius-Saba, Sint Barthelemy-Belgium-Luxembourg-Netherlands-Denmark-Iceland-Faroe Islands-Jersey Channel Islands-Cyprus-Turkey-Israel-Palest''
    
    \textit{Keywords:} econom, haiti, economi, bahama, economi
\end{enumerate}

\subsubsection{Hallucinated Responses}
\begin{enumerate}
    \item \textbf{Response H-1:} ``sierpina 1, 2022 Respond to the following questions with at least one reference from outside of your textbook. Your response should be at least two paragraphs long (not including references). What is an example of how you have used technology for learning or work purposes during this course? How has using that technology impacted you as a learner/worker?''
    
    \textit{Keywords:} use, learner, learn, textbook, technolog
    
    \item \textbf{Response H-2:} ``nobody is perfect. We all make mistakes, but it’s how we learn from them that matters most.” – Unknown “If you want something you never had before, then you have to do something you’ve never done before”- Unknown “The best way out of your problem is through your problem”. -Richard Gere “We are what our thoughts have made us; so take care about what you think. Words are secondary. Thoughts live; they travel far.” — Swami Vivekananda “I am not afraid of storms for I am learning to sail my ship.”—Louisa May Alcott “You can’t cross a chasm in two small jumps.”– David Foster Wallace “Everything has beauty, but not everyone sees it.”― Confucius “There will be obstacles. There will be doubters. And there will be haters. If you let those things control you, they win. Let go! Laugh at yourself, love who you are, embrace who you were meant to be, own everything you are, and move forward with passion!” – Will Smith “Whenever you find yourself on the side of the majority, it is time to pause and reflect.”—Mark Twain “Don’t tell me where good ideas come from. Tell me where bad ones go.”—Anonymous “A man must first believe he is worthy of happiness…then work like hell to achieve''
    
    \textit{Keywords:} swami, mistak, sail, obstacl, confuciu
\end{enumerate}

\subsection{Mathematical Formula}

\subsubsection{Kernel Density Estimation}
The Kernel Density Estimation (KDE) of a probability density function \( f(x) \) at a point \( x \), based on a dataset \( \{x_1, x_2, \dots, x_n\} \), is defined as:
\begin{equation}
    \hat{f}(x) = \frac{1}{n h} \sum_{i=1}^n K\left( \frac{x - x_i}{h} \right),
\end{equation}
where:
\begin{itemize}
    \item \( K(\cdot) \) is the kernel function, such as Gaussian, Epanechnikov, or Uniform.
    \item \( h > 0 \) is the bandwidth parameter that controls the smoothness of the density estimate.
    \item \( n \) is the number of data points in the dataset.
\end{itemize}

\newpage
\subsection{Training Results}
\subsubsection{$r=4$}

\begin{figure*}[h!]
    \centering
    \begin{subfigure}
        \centering
        \includegraphics[width=0.5\columnwidth]{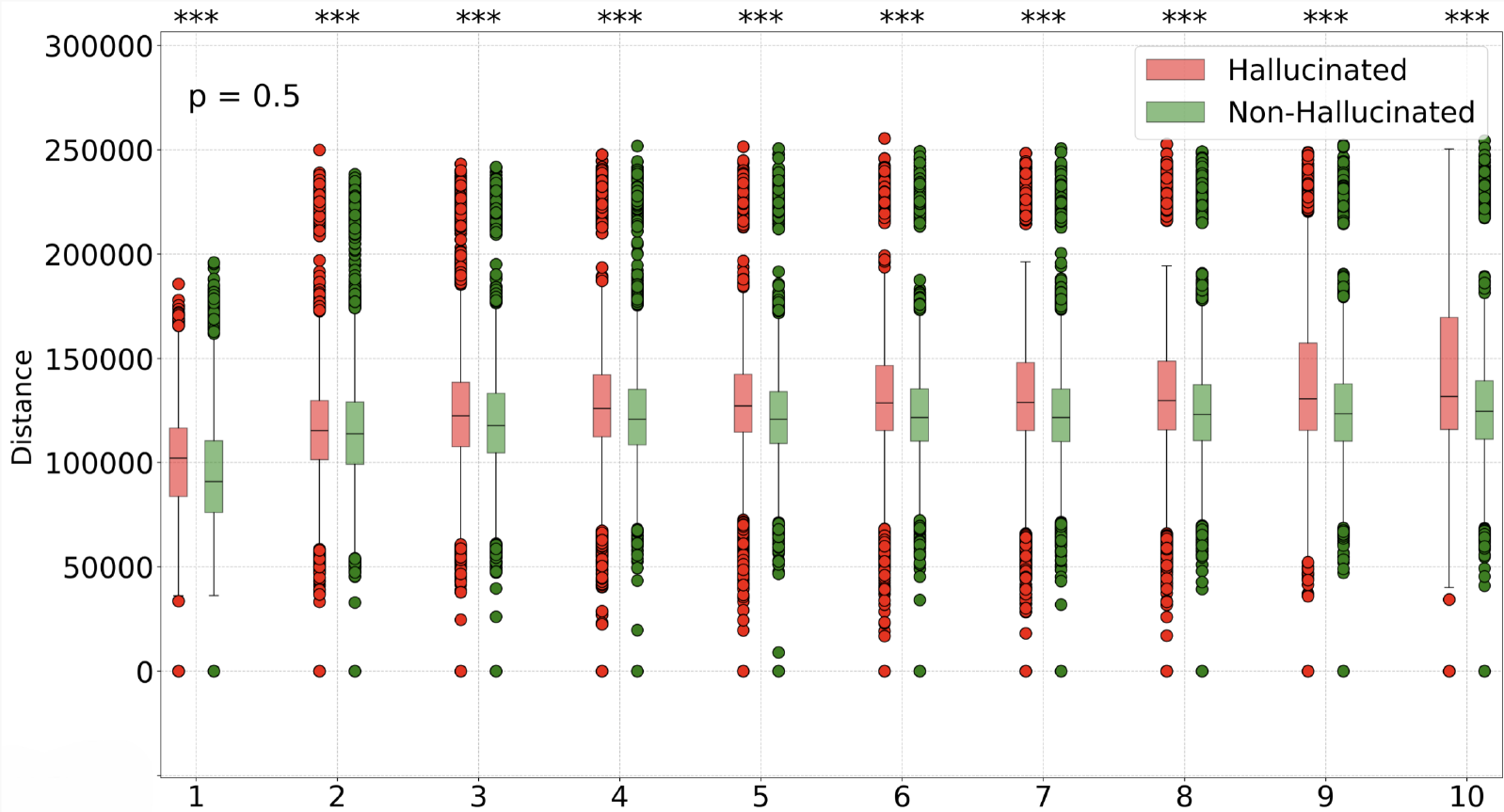}
    \end{subfigure}
    \hfill
    \begin{subfigure}
        \centering
        \includegraphics[width=0.49\columnwidth]{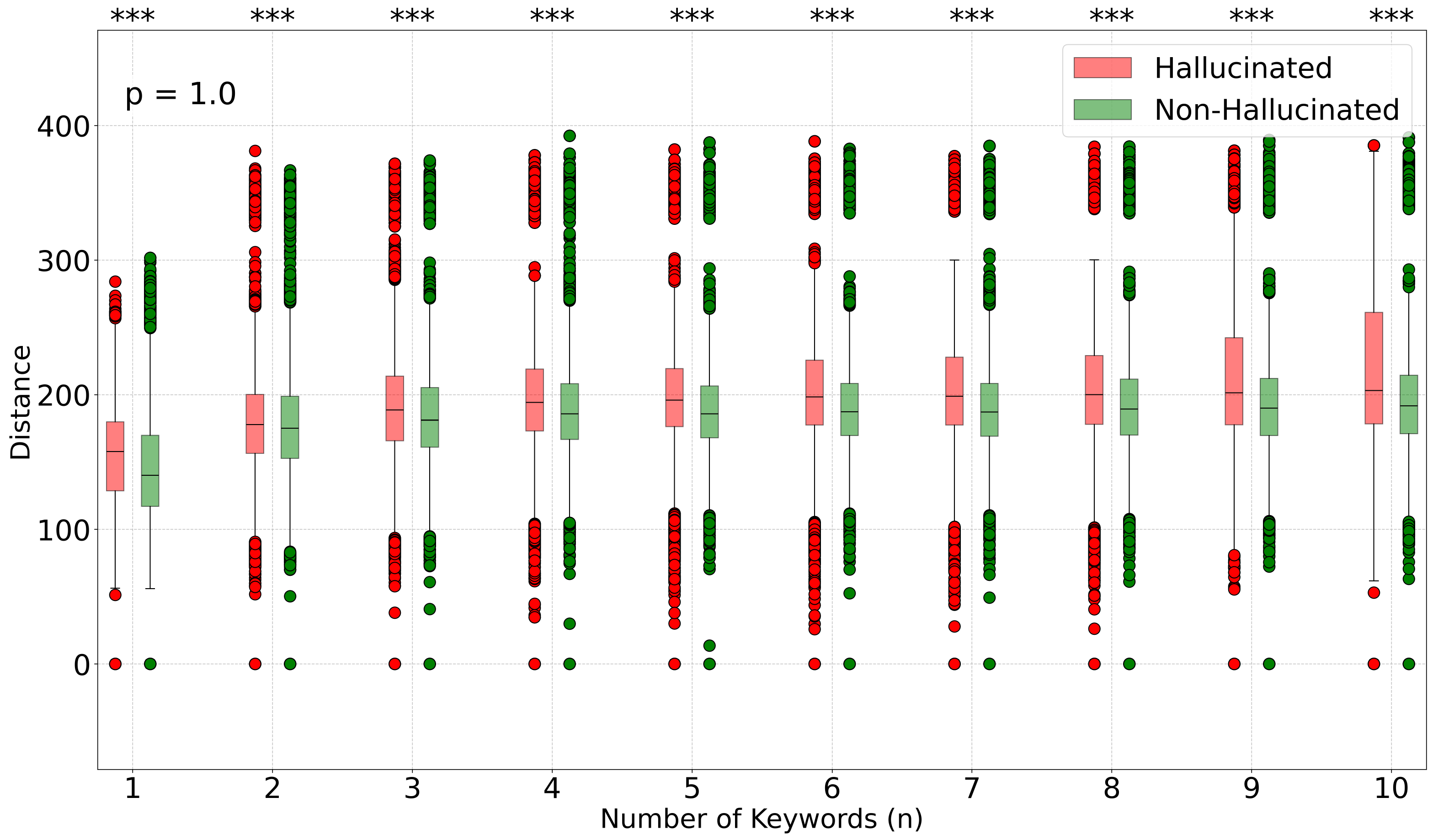}
    \end{subfigure}
    \begin{subfigure}
        \centering
        \includegraphics[width=0.5\columnwidth]{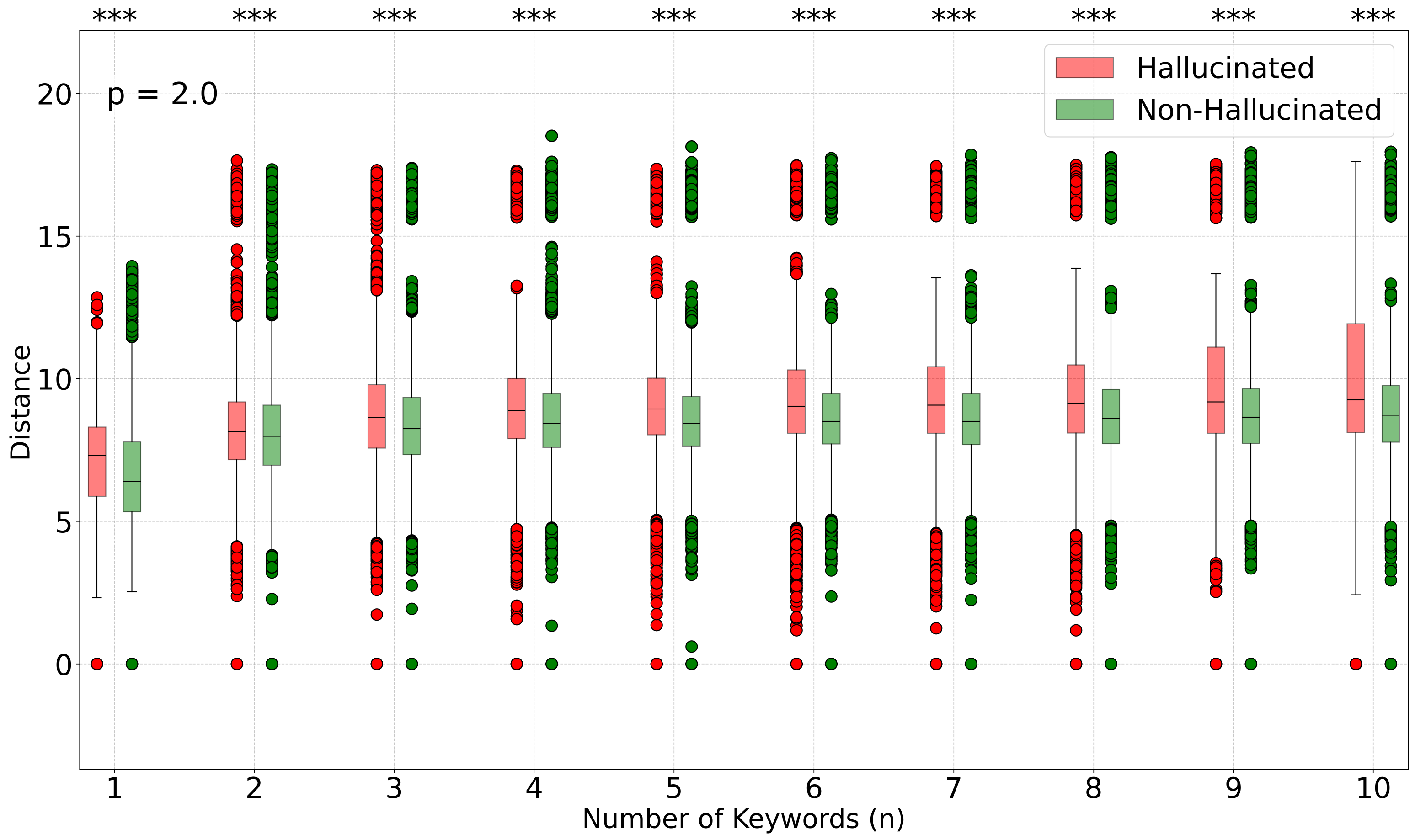}
    \end{subfigure}
    \caption{Training distributions of Minkowski distances with $r=4$.}
    \label{fig:r_4}
\end{figure*}

\renewcommand{\thetable}{\arabic{table}}
\begin{table}[H]
    \centering
    \begin{tabular}{||c|c|c|c|c|c|c|c|c|c|c|c||} 
     \hline
     & $n=1$ & $n=2$ & $n=3$ & $n=4$ & $n=5$ & $n=6$ & $n=7$ & $n=8$ & $n=9$ & $n=10$ \\ [0.5ex] 
     \hline\hline
    $p=0.5$ & 0.092 & 0.008 & 0.073 & 0.069 & 0.093 & 0.140 & 0.166 & 0.175 & 0.247 & 0.216 \\ 
     \hline
    $p=1.0$ & 0.094 & 0.008 & 0.078 & 0.071 & 0.104 & 0.143 & 0.165 & 0.172 & 0.254 & 0.212 \\
     \hline
    $p=2.0$ & 0.110 & 0.012 & 0.106 & 0.077 & 0.114 & 0.174 & 0.171 & 0.176 & 0.252 & 0.209 \\ 
     \hline
    \end{tabular}
    \caption{KL Divergence as a function of $n$ and $p$ for $r=4$}
    \label{tab:KL Table r=4}
\end{table}

\begin{table}[H]
    \centering
    \setlength{\tabcolsep}{3pt}
    \begin{tabular}{||c|c|c|c|c|c|c|c|c|c|c|c||} 
     \hline
     & $n=1$ & $n=2$ & $n=3$ & $n=4$ & $n=5$ & $n=6$ & $n=7$ & $n=8$ & $n=9$ & $n=10$ \\ [0.5ex] 
     \hline\hline
    $p=0.5$ & 11165.276 & 1544.044 & 4753.059 & 5241.775 & 6452.003 & 7106.216 & 7426.346 & 6965.591 & 7273.278 & 7323.110 \\ 
     \hline
    $p=1.0$ & 17.629 & 2.643 & 7.548 & 8.493 & 10.120 & 11.057 & 11.627 & 10.707 & 11.308 & 11.404 \\
     \hline
    $p=2.0$ & 0.909 & 0.160 & 0.389 & 0.437 & 0.500 & 0.520 & 0.563 & 0.528 & 0.537 & 0.532 \\ 
     \hline
    \end{tabular}
    \caption{Median Difference as a function of $n$ and $p$ for $r=4$}
    \label{tab:Median Table r=4}
\end{table}
\newpage

\subsubsection{$r=6$}

\begin{figure*}[h!]
    \centering
    \begin{subfigure}
        \centering
        \includegraphics[width=0.49\columnwidth]{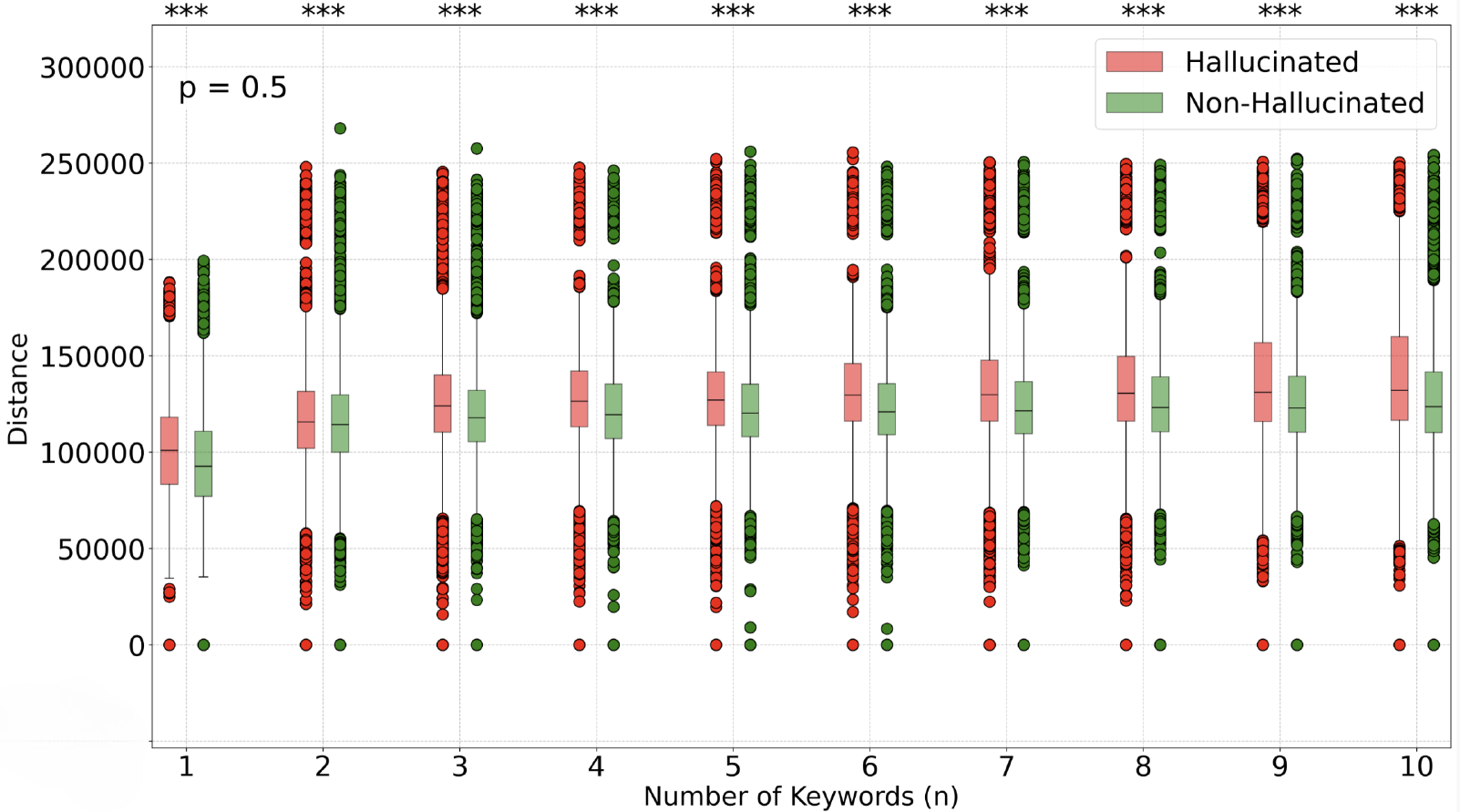}
    \end{subfigure}
    \hfill
    \begin{subfigure}
        \centering
        \includegraphics[width=0.49\columnwidth]{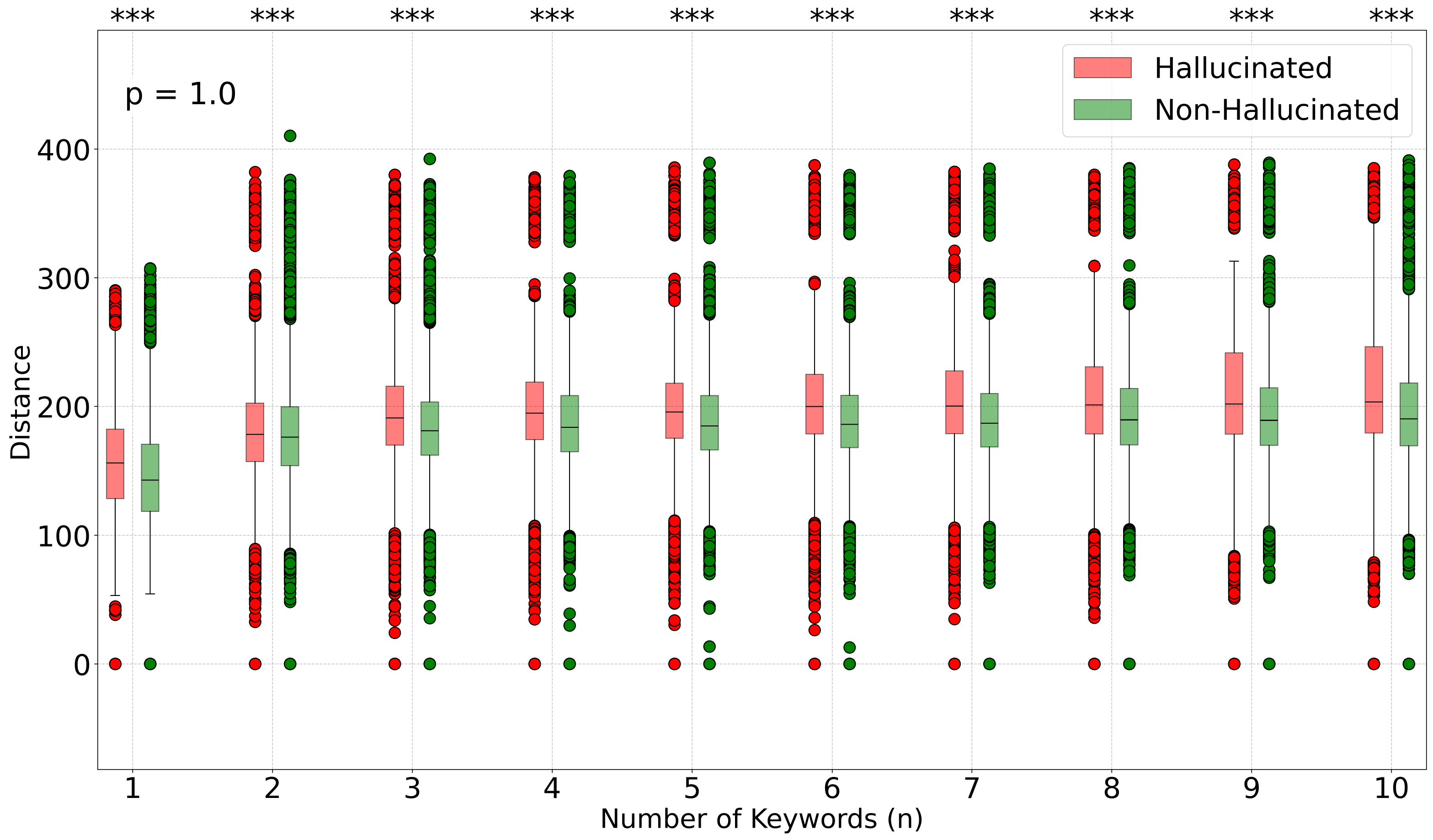}
    \end{subfigure}
    \begin{subfigure}
        \centering
        \includegraphics[width=0.5\columnwidth]{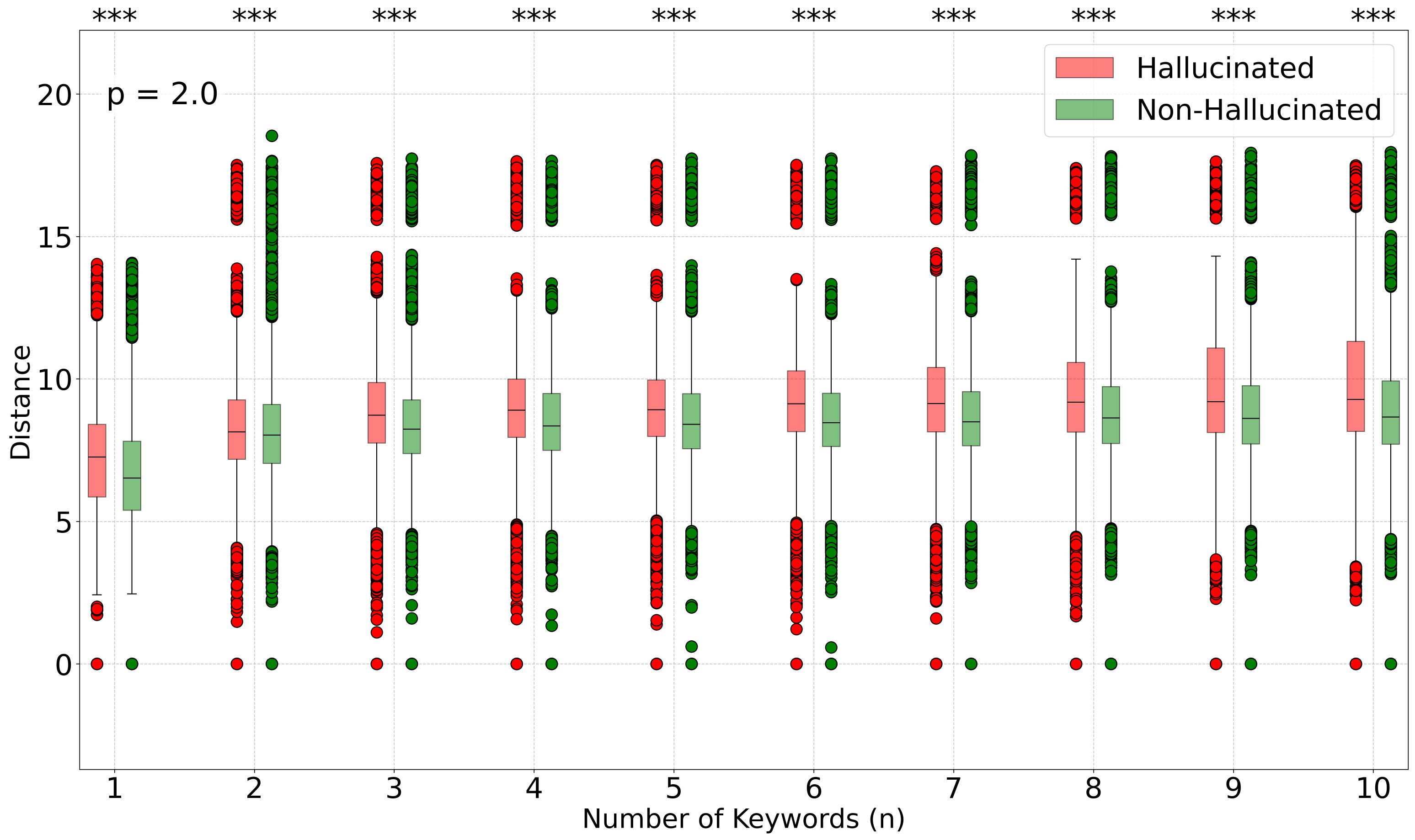}
    \end{subfigure}
    \caption{Training distributions of Minkowski distances with $r=6$.}
    \label{fig:r_6}
\end{figure*}

\begin{table}[H]
    \centering
    \begin{tabular}{||c|c|c|c|c|c|c|c|c|c|c|c||} 
     \hline
     & $n=1$ & $n=2$ & $n=3$ & $n=4$ & $n=5$ & $n=6$ & $n=7$ & $n=8$ & $n=9$ & $n=10$ \\ [0.5ex] 
     \hline\hline
    $p=0.5$ & 0.055 & 0.015 & 0.066 & 0.071 & 0.075 & 0.109 & 0.096 & 0.076 & 0.097 & 0.095 \\ 
     \hline
    $p=1.0$ & 0.060 & 0.017 & 0.065 & 0.071 & 0.078 & 0.105 & 0.096 & 0.081 & 0.096 & 0.097 \\
     \hline
    $p=2.0$ & 0.075 & 0.021 & 0.065 & 0.082 & 0.083 & 0.114 & 0.099 & 0.082 & 0.099 & 0.098 \\ 
     \hline
    \end{tabular}
    \caption{KL Divergence as a function of $n$ and $p$ for $r=6$}
    \label{tab:KL Table r=6}
\end{table}

\begin{table}[H]
    \centering
    \setlength{\tabcolsep}{3pt}
    \begin{tabular}{||c|c|c|c|c|c|c|c|c|c|c|c||} 
     \hline
     & $n=1$ & $n=2$ & $n=3$ & $n=4$ & $n=5$ & $n=6$ & $n=7$ & $n=8$ & $n=9$ & $n=10$ \\ [0.5ex] 
     \hline\hline
    $p=0.5$ & 8195.671 & 1337.687 & 6245.095 & 6925.253 & 6795.699 & 8700.962 & 8401.756 & 7378.489 & 8187.909 & 8461.692 \\ 
     \hline
    $p=1.0$ & 13.250 & 2.074 & 9.787 & 10.887 & 10.544 & 13.598 & 13.155 & 11.598 & 12.638 & 13.091 \\
     \hline
    $p=2.0$ & 0.740 & 0.112 & 0.479 & 0.543 & 0.504 & 0.657 & 0.630 & 0.559 & 0.588 & 0.612 \\ 
     \hline
    \end{tabular}
    \caption{Median Difference as a function of $n$ and $p$ for $r=6$}
    \label{tab:Median Table r=6}
\end{table}

\newpage

\subsubsection{$r=8$}

\begin{figure*}[h!]
    \centering
    \begin{subfigure}
        \centering
        \includegraphics[width=0.49\columnwidth]{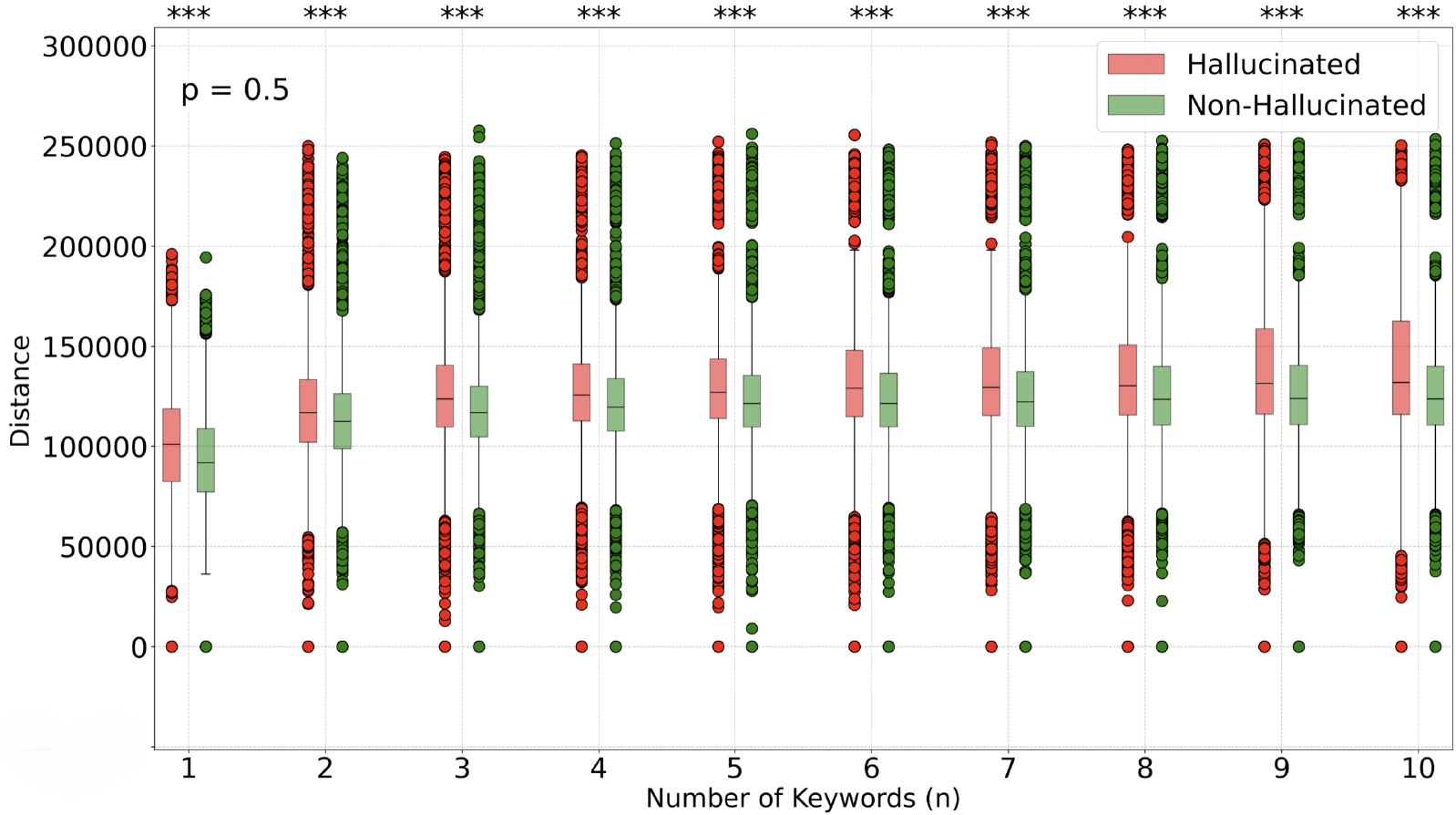}
    \end{subfigure}
    \hfill
    \begin{subfigure}
        \centering
        \includegraphics[width=0.49\columnwidth]{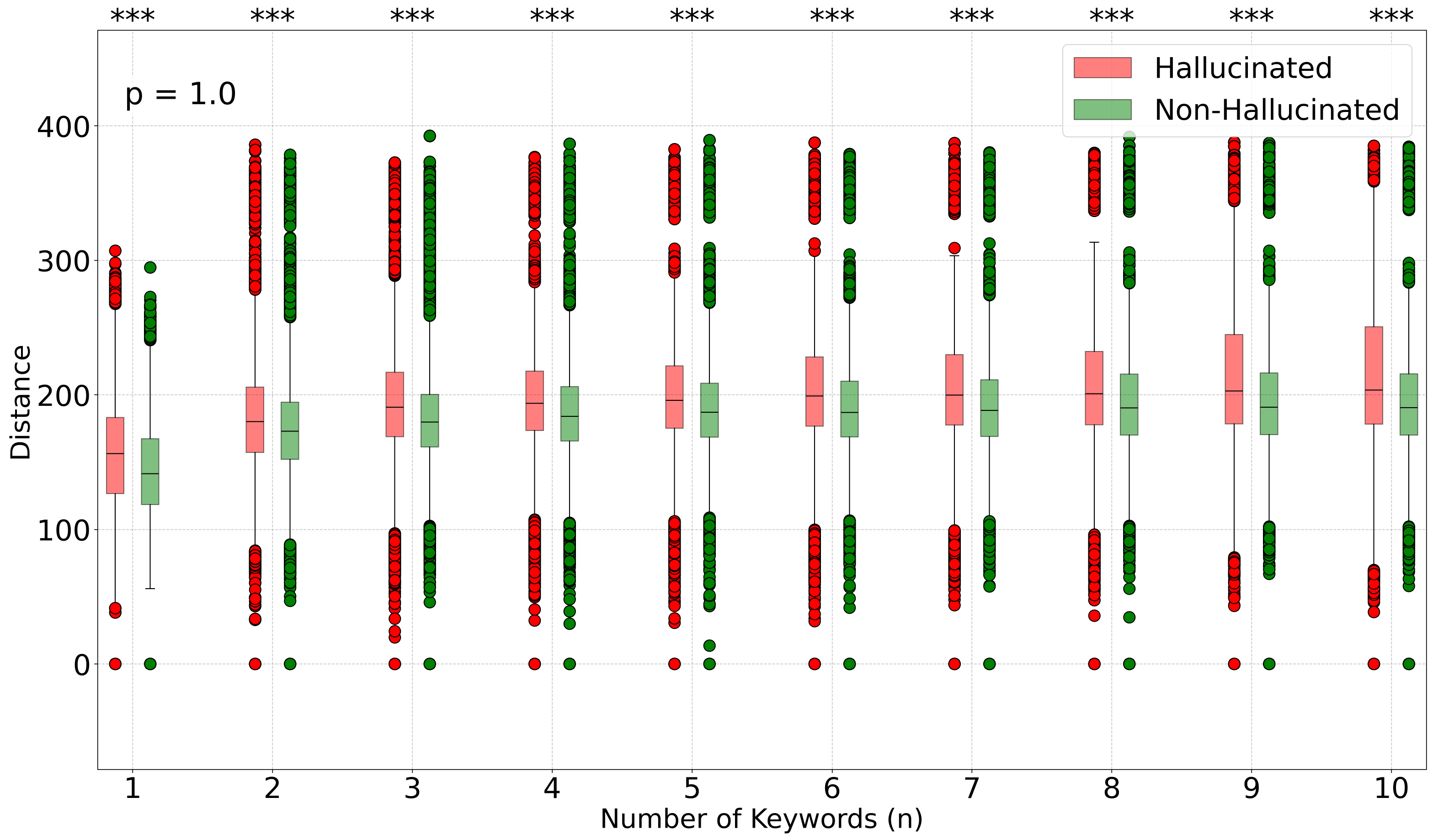}
    \end{subfigure}
    \begin{subfigure}
        \centering
        \includegraphics[width=0.5\columnwidth]{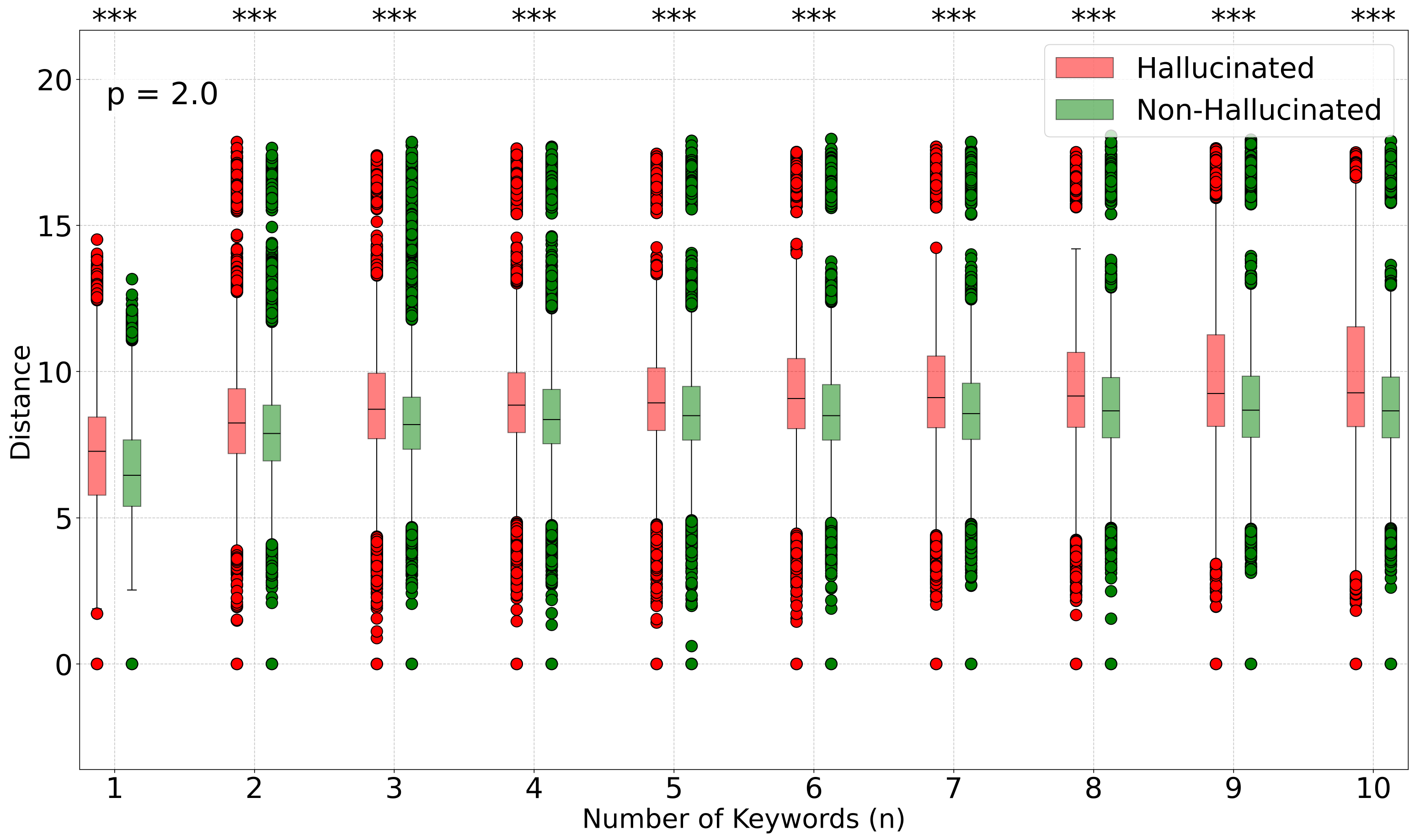}
    \end{subfigure}
    \caption{Training distributions of Minkowski distances with $r=8$.}
    \label{fig:r_8}
\end{figure*}

\begin{table}[H]
    \centering
    \begin{tabular}{||c|c|c|c|c|c|c|c|c|c|c|c||} 
     \hline
     & $n=1$ & $n=2$ & $n=3$ & $n=4$ & $n=5$ & $n=6$ & $n=7$ & $n=8$ & $n=9$ & $n=10$ \\ [0.5ex] 
     \hline\hline
    $p=0.5$ & 0.082 & 0.035 & 0.099 & 0.056 & 0.056 & 0.085 & 0.078 & 0.061 & 0.082 & 0.095 \\ 
     \hline
    $p=1.0$ & 0.092 & 0.036 & 0.101 & 0.058 & 0.057 & 0.088 & 0.079 & 0.060 & 0.081 & 0.100 \\
     \hline
    $p=2.0$ & 0.121 & 0.040 & 0.106 & 0.062 & 0.062 & 0.087 & 0.080 & 0.062 & 0.085 & 0.102 \\ 
     \hline
    \end{tabular}
    \caption{KL Divergence as a function of $n$ and $p$ for $r=8$}
    \label{tab:KL Table r=8}
\end{table}

\begin{table}[H]
    \centering
    \setlength{\tabcolsep}{3pt}
    \begin{tabular}{||c|c|c|c|c|c|c|c|c|c|c|c||} 
     \hline
     & $n=1$ & $n=2$ & $n=3$ & $n=4$ & $n=5$ & $n=6$ & $n=7$ & $n=8$ & $n=9$ & $n=10$ \\ [0.5ex] 
     \hline\hline
    $p=0.5$ & 9107.327 & 4464.322 & 6920.413 & 6210.219 & 5595.131 & 7770.864 & 7303.950 & 6682.462 & 7630.817 & 8333.385 \\ 
     \hline
    $p=1.0$ & 14.875 & 7.161 & 10.868 & 9.804 & 8.882 & 12.190 & 11.434 & 10.410 & 11.981 & 13.021 \\
     \hline
    $p=2.0$ & 0.823 & 0.362 & 0.519 & 0.486 & 0.444 & 0.588 & 0.545 & 0.504 & 0.570 & 0.615 \\ 
     \hline
    \end{tabular}
    \caption{Median Difference as a function of $n$ and $p$ for $r=8$}
    \label{tab:Median Table r=8}
\end{table}

\newpage

\subsubsection{$r=10$}

\begin{figure*}[h!]
    \centering
    \begin{subfigure}
        \centering
        \includegraphics[width=0.49\columnwidth]{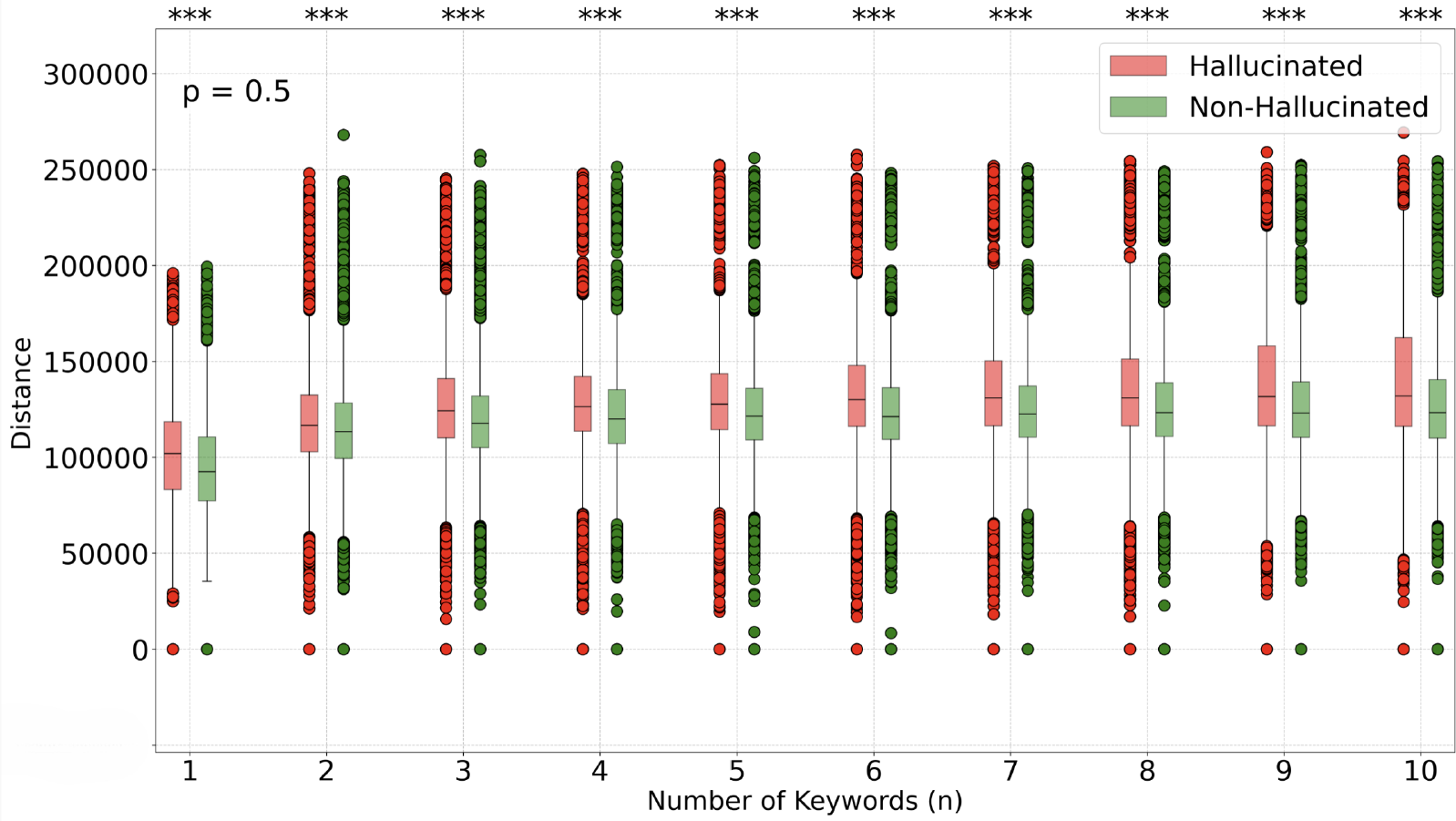}
    \end{subfigure}
    \hfill
    \begin{subfigure}
        \centering
        \includegraphics[width=0.49\columnwidth]{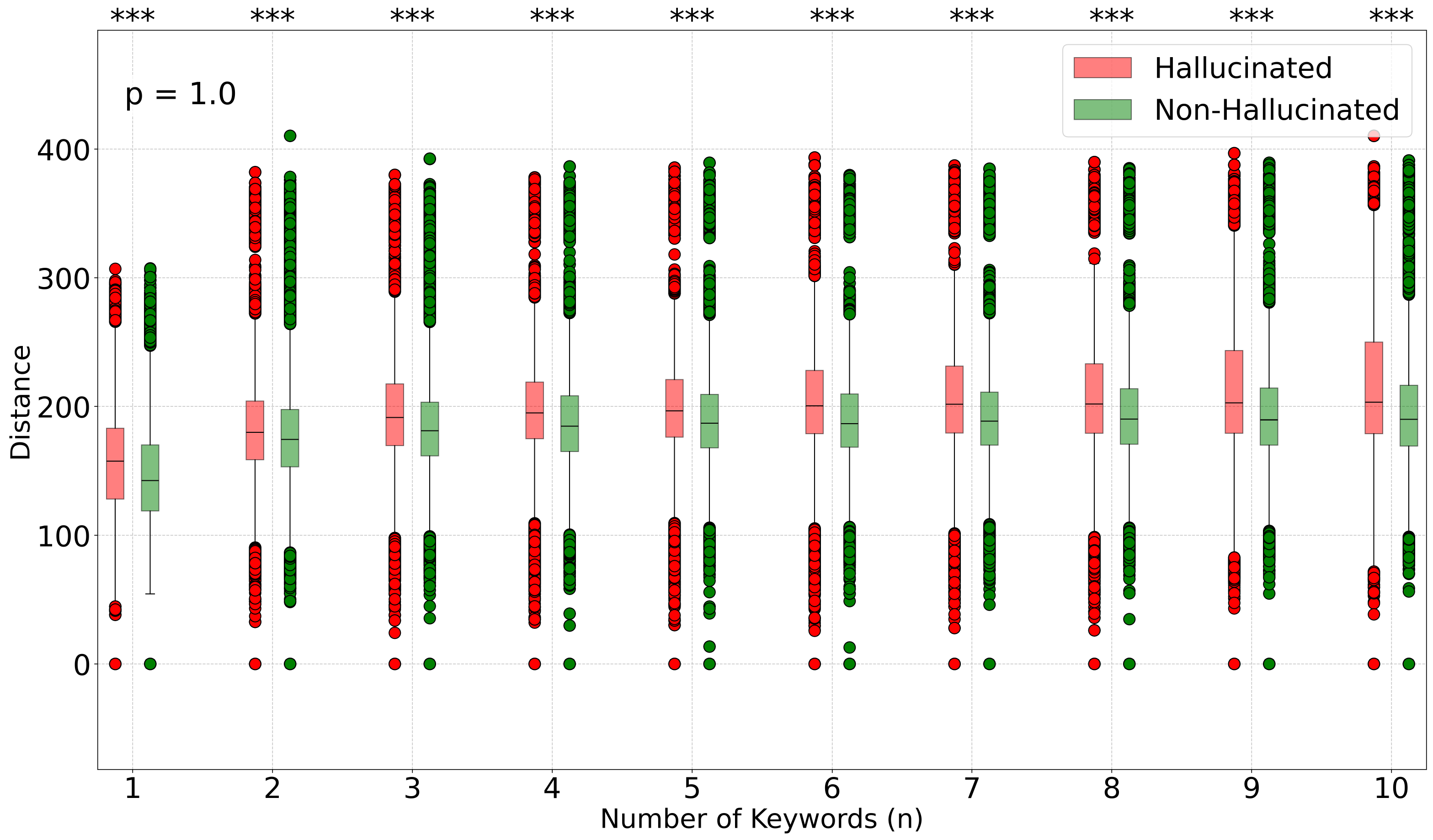}
    \end{subfigure}
    \begin{subfigure}
        \centering
        \includegraphics[width=0.5\columnwidth]{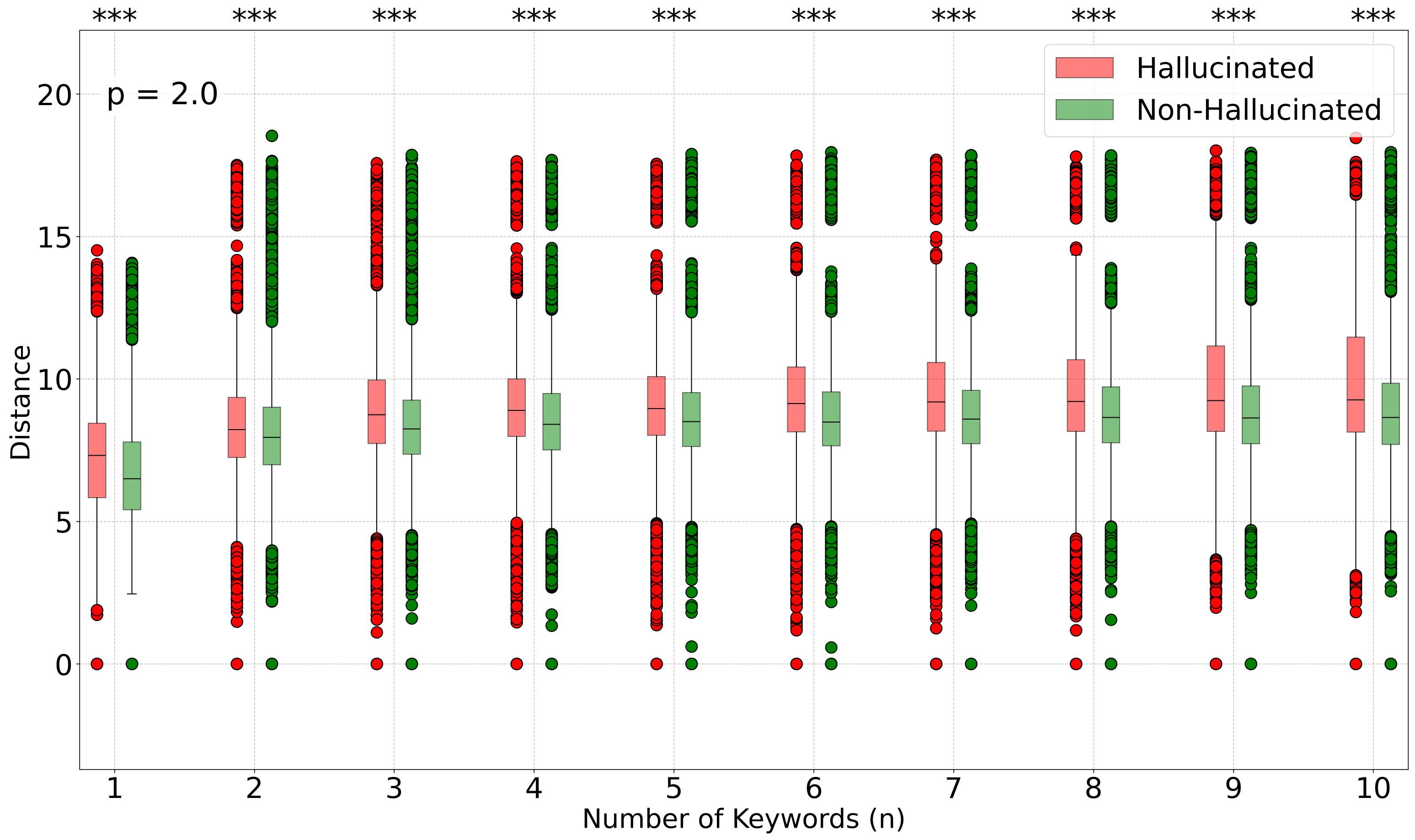}
    \end{subfigure}
    \caption{Training distributions of Minkowski distances with $r=10$.}
    \label{fig:r_10}
\end{figure*}

\begin{table}[H]
    \centering
    \begin{tabular}{||c|c|c|c|c|c|c|c|c|c|c|c||} 
     \hline
     & $n=1$ & $n=2$ & $n=3$ & $n=4$ & $n=5$ & $n=6$ & $n=7$ & $n=8$ & $n=9$ & $n=10$ \\ [0.5ex] 
     \hline\hline
    $p=0.5$ & 0.073 & 0.024 & 0.080 & 0.070 & 0.067 & 0.116 & 0.101 & 0.092 & 0.112 & 0.110 \\ 
     \hline
    $p=1.0$ & 0.078 & 0.025 & 0.081 & 0.071 & 0.068 & 0.125 & 0.101 & 0.085 & 0.109 & 0.110 \\
     \hline
    $p=2.0$ & 0.097 & 0.029 & 0.083 & 0.072 & 0.071 & 0.116 & 0.101 & 0.090 & 0.109 & 0.113 \\ 
     \hline
    \end{tabular}
    \caption{KL Divergence as a function of $n$ and $p$ for $r=10$}
    \label{tab:KL Table r=10}
\end{table}

\begin{table}[H]
    \centering
    \setlength{\tabcolsep}{3pt}
    \begin{tabular}{||c|c|c|c|c|c|c|c|c|c|c|c||} 
     \hline
     & $n=1$ & $n=2$ & $n=3$ & $n=4$ & $n=5$ & $n=6$ & $n=7$ & $n=8$ & $n=9$ & $n=10$ \\ [0.5ex] 
     \hline\hline
    $p=0.5$ & 9468.679 & 3465.374 & 6482.022 & 6411.284 & 6104.622 & 8736.044 & 8311.159 & 7560.490 & 8531.711 & 8624.816 \\ 
     \hline
    $p=1.0$ & 15.171 & 5.526 & 10.186 & 9.995 & 9.510 & 13.574 & 12.887 & 11.703 & 13.181 & 13.302 \\
     \hline
    $p=2.0$ & 0.822 & 0.270 & 0.488 & 0.484 & 0.451 & 0.645 & 0.606 & 0.559 & 0.614 & 0.617 \\ 
     \hline
    \end{tabular}
    \caption{Median Difference as a function of $n$ and $p$ for $r=10$}
    \label{tab:Median Table r=10}
\end{table}

\newpage
\subsubsection{$r=12$}

\begin{figure*}[h!]
    \centering
    \begin{subfigure}
        \centering
        \includegraphics[width=0.49\columnwidth]{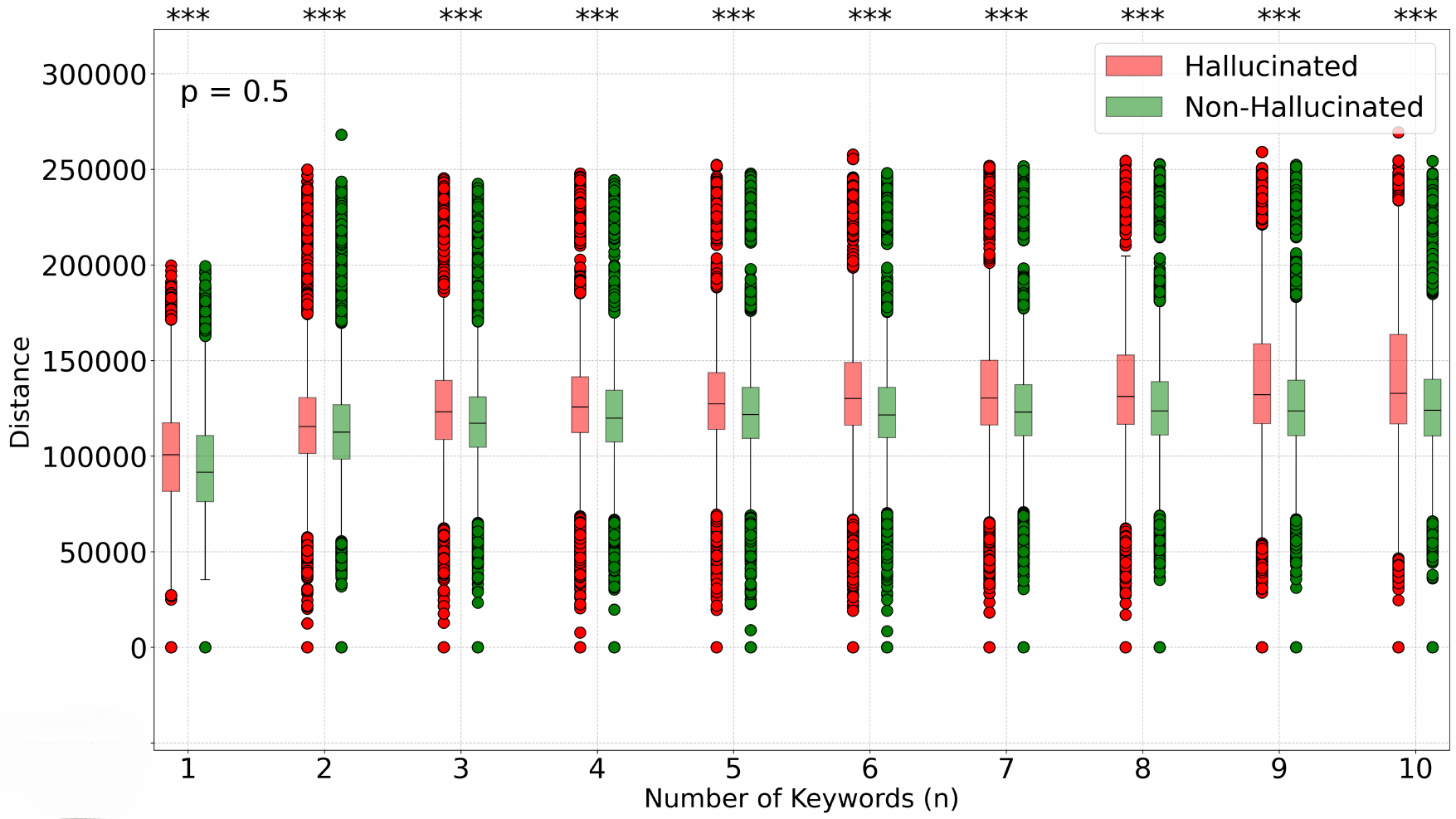}
    \end{subfigure}
    \hfill
    \begin{subfigure}
        \centering
        \includegraphics[width=0.49\columnwidth]{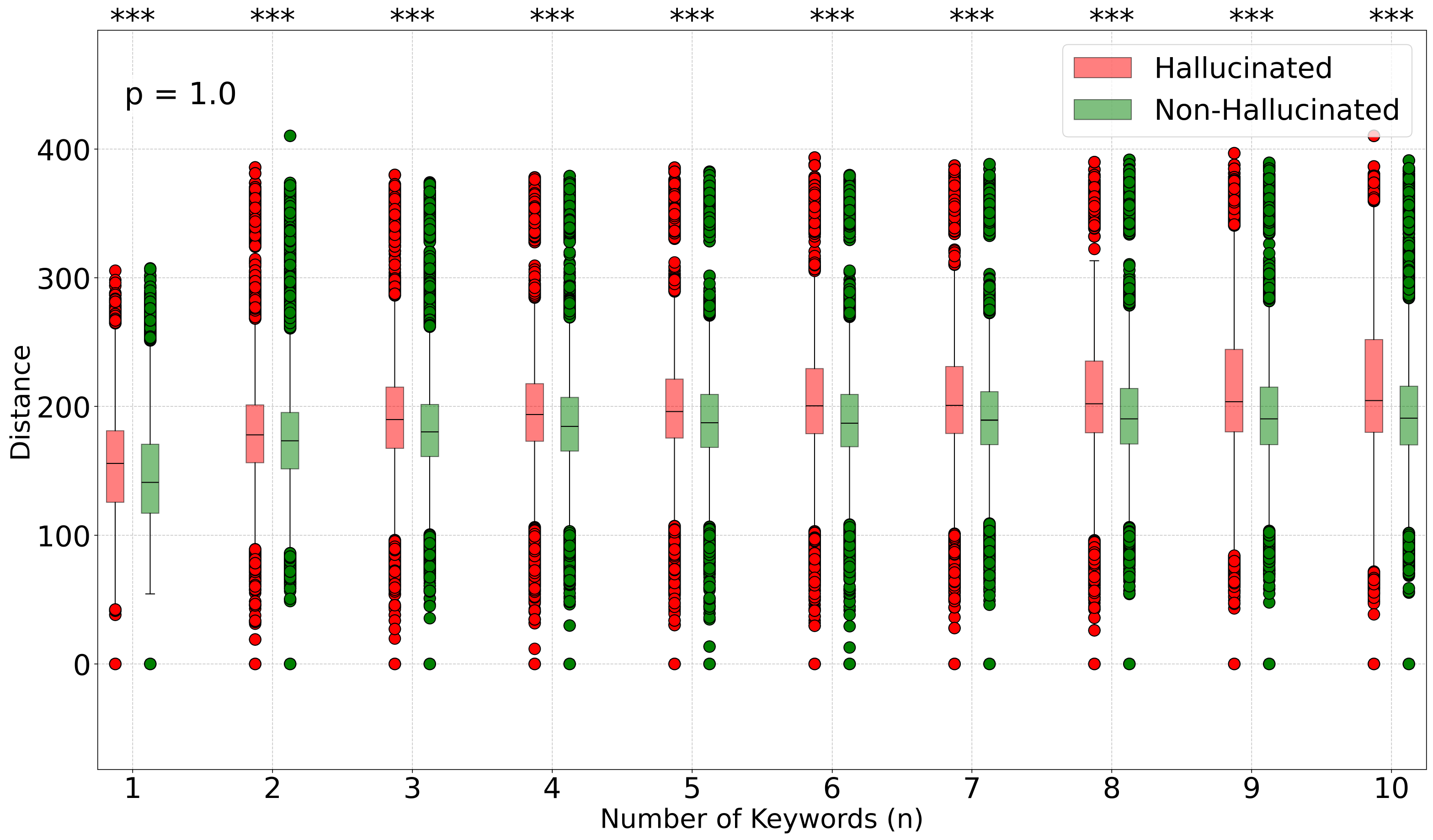}
    \end{subfigure}
    \begin{subfigure}
        \centering
        \includegraphics[width=0.5\columnwidth]{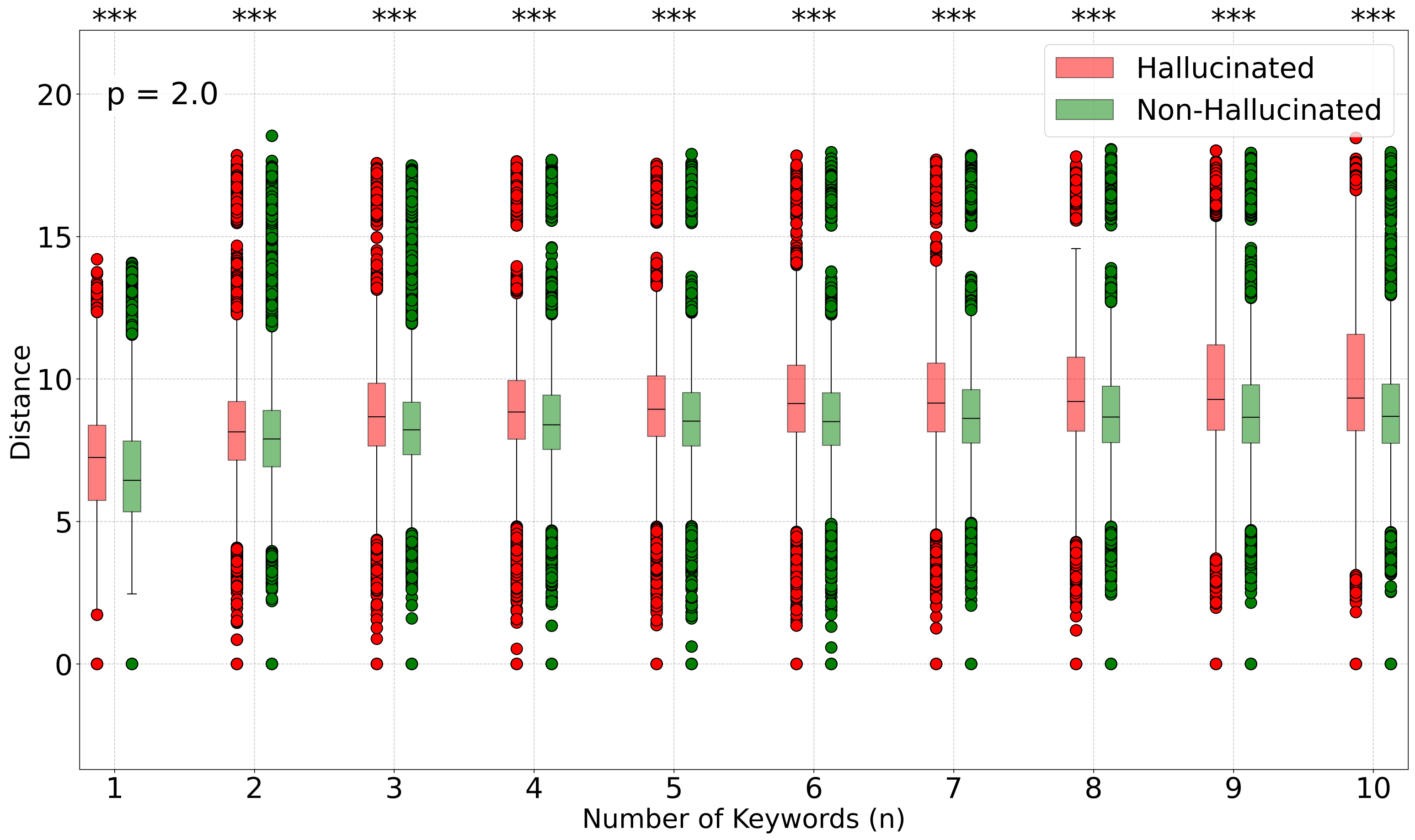}
    \end{subfigure}
    \caption{Training distributions of Minkowski distances with $r=12$.}
    \label{fig:r_12}
\end{figure*}

\begin{table}[H]
    \centering
    \begin{tabular}{||c|c|c|c|c|c|c|c|c|c|c|c||} 
     \hline
     & $n=1$ & $n=2$ & $n=3$ & $n=4$ & $n=5$ & $n=6$ & $n=7$ & $n=8$ & $n=9$ & $n=10$ \\ [0.5ex] 
     \hline\hline
    $p=0.5$ & 0.056 & 0.016 & 0.083 & 0.057 & 0.067 & 0.120 & 0.098 & 0.097 & 0.109 & 0.114 \\ 
     \hline
    $p=1.0$ & 0.062 & 0.018 & 0.085 & 0.057 & 0.066 & 0.124 & 0.099 & 0.094 & 0.108 & 0.118 \\
     \hline
    $p=2.0$ & 0.075 & 0.022 & 0.086 & 0.065 & 0.069 & 0.116 & 0.097 & 0.094 & 0.106 & 0.119 \\ 
     \hline
    \end{tabular}
    \caption{KL Divergence as a function of $n$ and $p$ for $r=12$}
    \label{tab:KL Table r=12}
\end{table}

\begin{table}[H]
    \centering
    \setlength{\tabcolsep}{3pt}
    \begin{tabular}{||c|c|c|c|c|c|c|c|c|c|c|c||} 
     \hline
     & $n=1$ & $n=2$ & $n=3$ & $n=4$ & $n=5$ & $n=6$ & $n=7$ & $n=8$ & $n=9$ & $n=10$ \\ [0.5ex] 
     \hline\hline
    $p=0.5$ & 9125.015 & 2890.849 & 5973.321 & 5747.743 & 5526.221 & 8614.343 & 7418.399 & 7574.615 & 8594.014 & 8873.069 \\ 
     \hline
    $p=1.0$ & 14.784 & 4.691 & 9.399 & 9.001 & 8.603 & 13.346 & 11.460 & 11.701 & 13.262 & 13.733 \\
     \hline
    $p=2.0$ & 0.800 & 0.248 & 0.452 & 0.439 & 0.408 & 0.626 & 0.533 & 0.545 & 0.624 & 0.641 \\ 
     \hline
    \end{tabular}
    \caption{Median Difference as a function of $n$ and $p$ for $r=12$}
    \label{tab:Median Table r=12}
\end{table}

\newpage
\subsubsection{$r=14$}

\begin{figure*}[h!]
    \centering
    \begin{subfigure}
        \centering
        \includegraphics[width=0.49\columnwidth]{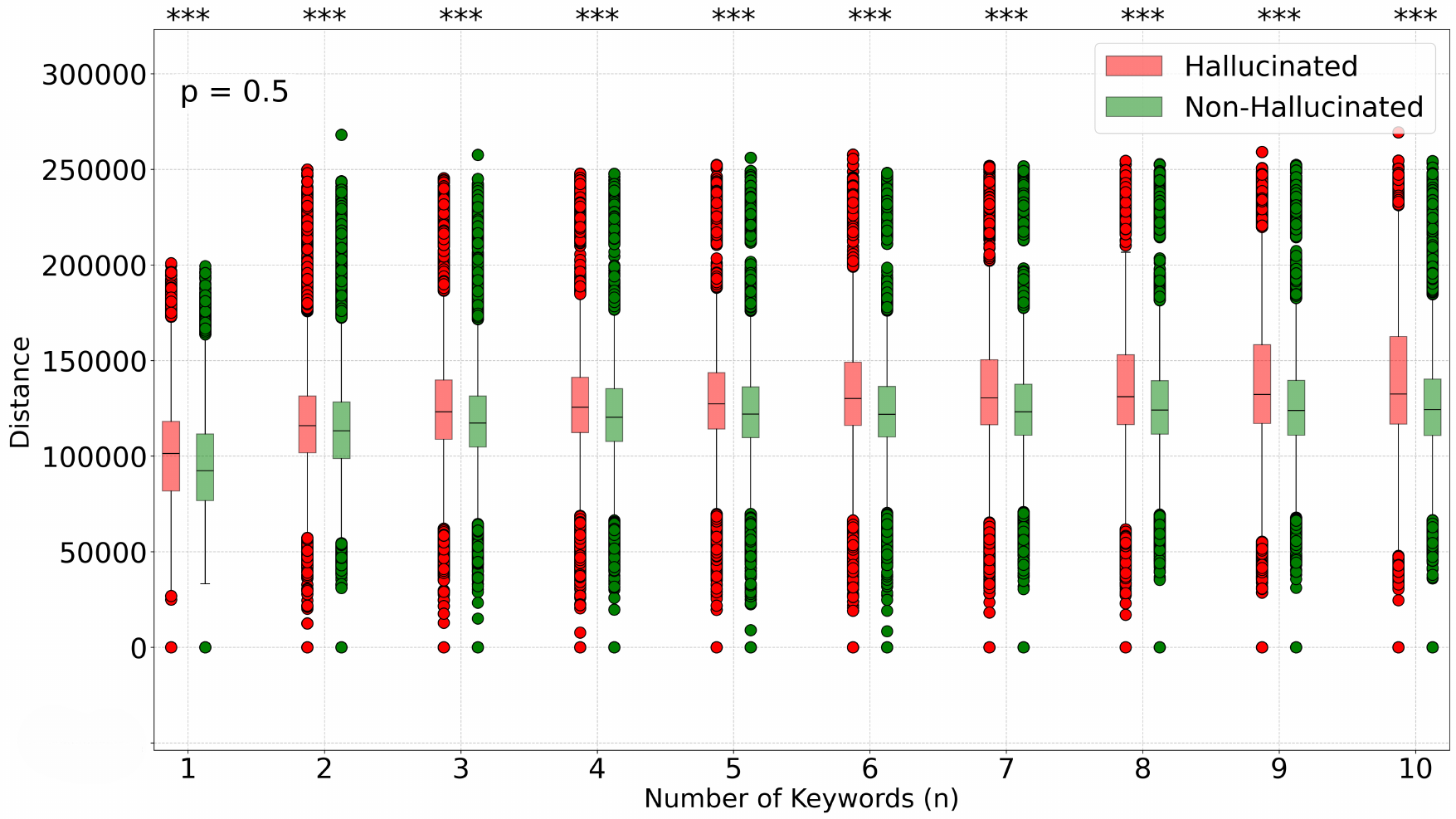}
    \end{subfigure}
    \hfill
    \begin{subfigure}
        \centering
        \includegraphics[width=0.49\columnwidth]{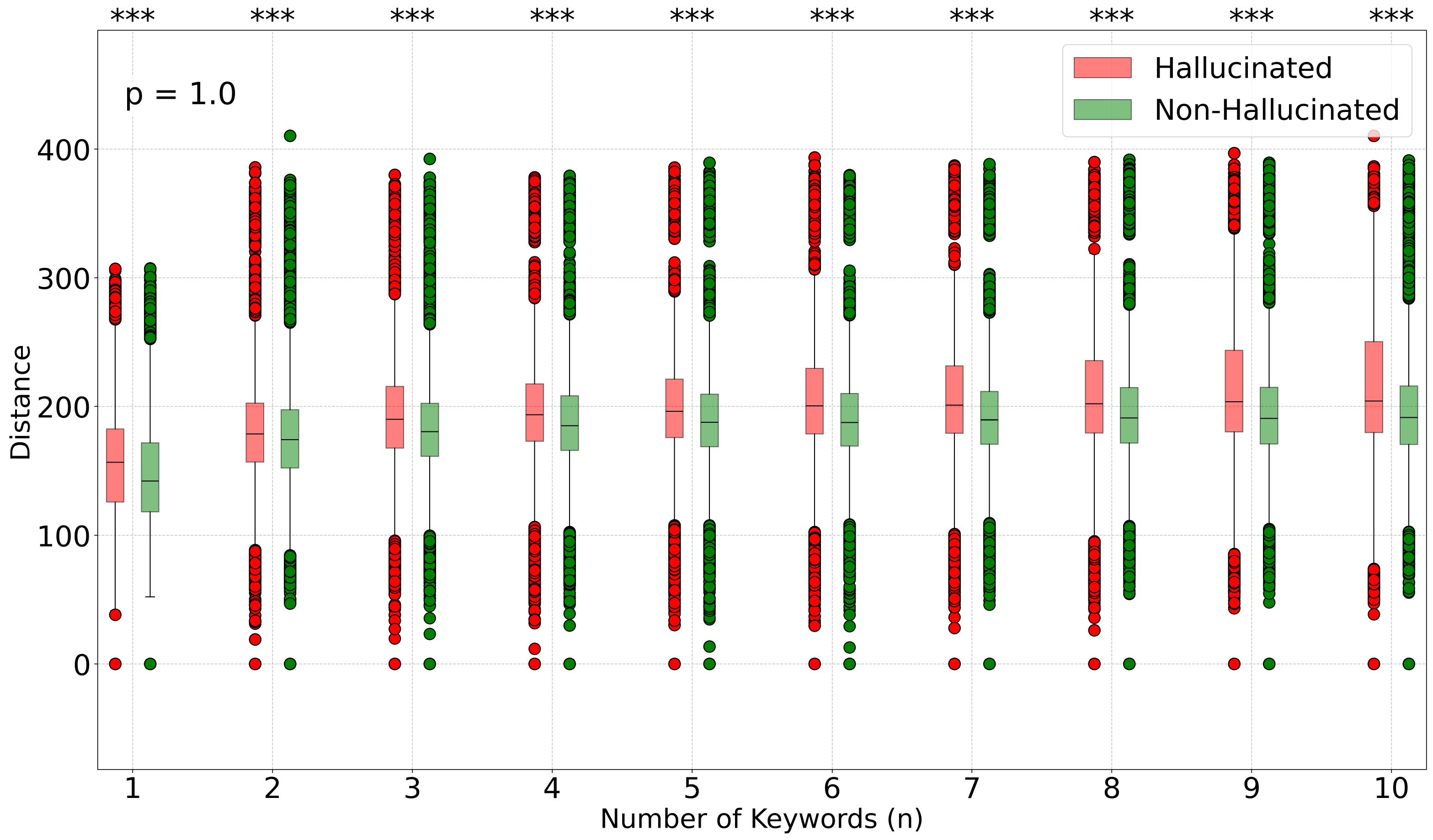}
    \end{subfigure}
    \begin{subfigure}
        \centering
        \includegraphics[width=0.5\columnwidth]{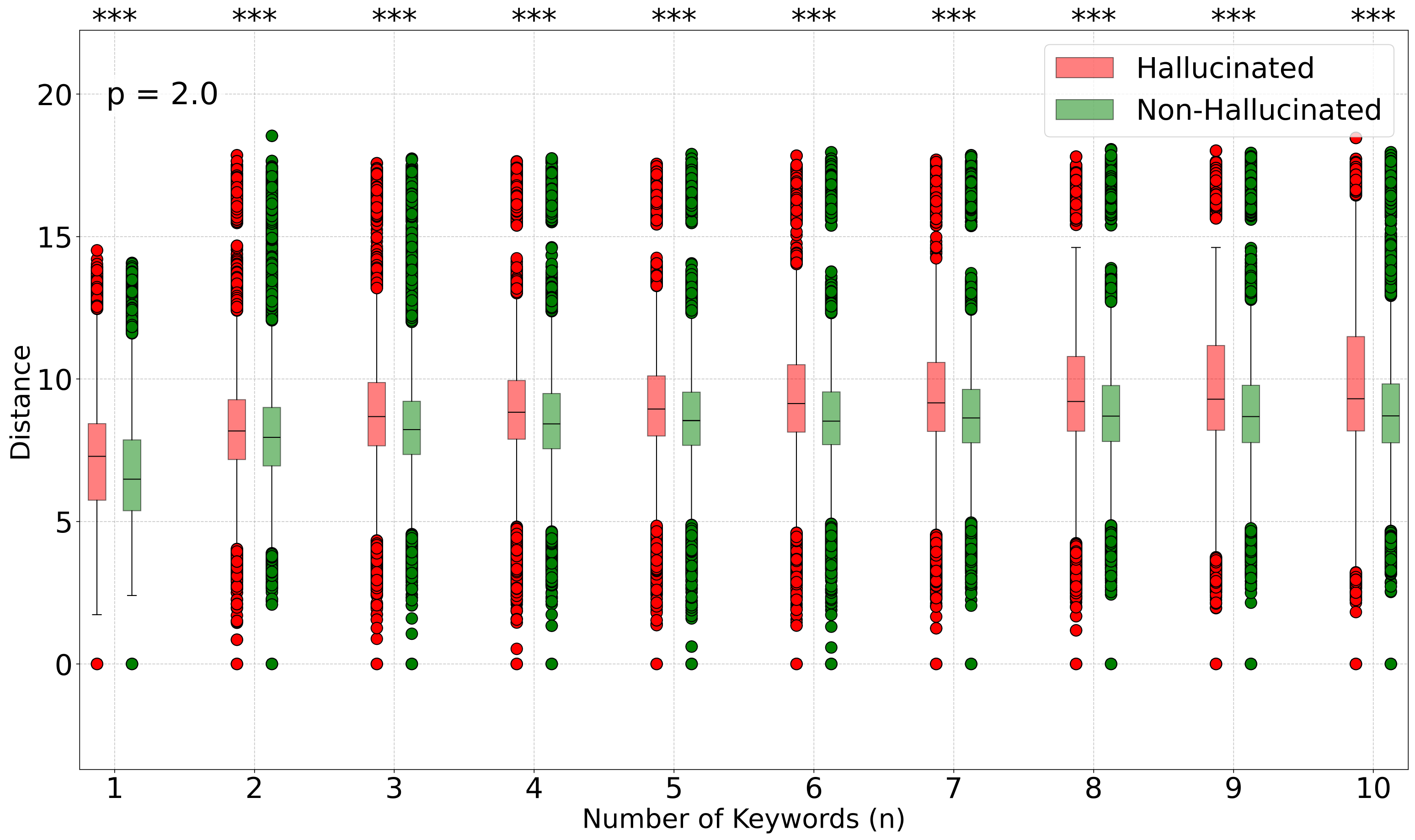}
    \end{subfigure}
    \caption{Training distributions of Minkowski distances with $r=14$.}
    \label{fig:r_14}
\end{figure*}

\begin{table}[H]
    \centering
    \begin{tabular}{||c|c|c|c|c|c|c|c|c|c|c|c||} 
     \hline
     & $n=1$ & $n=2$ & $n=3$ & $n=4$ & $n=5$ & $n=6$ & $n=7$ & $n=8$ & $n=9$ & $n=10$ \\ [0.5ex] 
     \hline\hline
    $p=0.5$ & 0.052 & 0.013 & 0.072 & 0.05 & 0.062 & 0.117 & 0.100 & 0.096 & 0.110 & 0.110 \\ 
     \hline
    $p=1.0$ & 0.056 & 0.015 & 0.074 & 0.05 & 0.062 & 0.121 & 0.101 & 0.092 & 0.108 & 0.113 \\
     \hline
    $p=2.0$ & 0.073 & 0.017 & 0.077 & 0.05 & 0.064 & 0.116 & 0.101 & 0.093 & 0.109 & 0.116 \\ 
     \hline
    \end{tabular}
    \caption{KL Divergence as a function of $n$ and $p$ for $r=14$}
    \label{tab:KL Table r=14}
\end{table}

\begin{table}[H]
    \centering
    \setlength{\tabcolsep}{3pt}
    \begin{tabular}{||c|c|c|c|c|c|c|c|c|c|c|c||} 
     \hline
     & $n=1$ & $n=2$ & $n=3$ & $n=4$ & $n=5$ & $n=6$ & $n=7$ & $n=8$ & $n=9$ & $n=10$ \\ [0.5ex] 
     \hline\hline
    $p=0.5$ & 9008.420 & 2671.746 & 5923.855 & 5212.144 & 5299.904 & 8205.322 & 7332.568 & 7001.040 & 8331.338 & 8261.098 \\ 
     \hline
    $p=1.0$ & 14.606 & 4.343 & 9.338 & 8.227 & 8.282 & 12.735 & 11.367 & 10.861 & 12.902 & 12.790 \\
     \hline
    $p=2.0$ & 0.797 & 0.226 & 0.453 & 0.403 & 0.400 & 0.604 & 0.534 & 0.515 & 0.609 & 0.598 \\ 
     \hline
    \end{tabular}
    \caption{Median Difference as a function of $n$ and $p$ for $r=14$}
    \label{tab:Median Table r=14}
\end{table}

\newpage
\subsubsection{$r=16$}

The train boxplots with $r=16$ are already reported in the paper as Figure \ref{fig:distance_distribution}.

\begin{table}[H]
    \centering
    \begin{tabular}{||c|c|c|c|c|c|c|c|c|c|c|c||} 
     \hline
     & $n=1$ & $n=2$ & $n=3$ & $n=4$ & $n=5$ & $n=6$ & $n=7$ & $n=8$ & $n=9$ & $n=10$ \\ [0.5ex] 
     \hline\hline
    $p=0.5$ & 0.048 & 0.013 & 0.071 & 0.056 & 0.069 & 0.121 & 0.109 & 0.109 & 0.133 & 0.134 \\ 
     \hline
    $p=1.0$ & 0.052 & 0.014 & 0.072 & 0.055 & 0.070 & 0.126 & 0.110 & 0.105 & 0.131 & 0.136 \\
     \hline
    $p=2.0$ & 0.068 & 0.017 & 0.076 & 0.056 & 0.072 & 0.121 & 0.109 & 0.106 & 0.132 & 0.138 \\ 
     \hline
    \end{tabular}
    \caption{KL Divergence as a function of $n$ and $p$ for $r=16$}
    \label{tab:KL Table r=16}
\end{table}

\begin{table}[H]
    \centering
    \setlength{\tabcolsep}{3pt}
    \begin{tabular}{||c|c|c|c|c|c|c|c|c|c|c|c||} 
     \hline
     & $n=1$ & $n=2$ & $n=3$ & $n=4$ & $n=5$ & $n=6$ & $n=7$ & $n=8$ & $n=9$ & $n=10$ \\ [0.5ex] 
     \hline\hline
    $p=0.5$ & 8842.034 & 2532.987 & 5651.462 & 5232.917 & 5270.153 & 8124.984 & 7499.443 & 7211.185 & 8748.346 & 8844.063 \\ 
     \hline
    $p=1.0$ & 14.129 & 4.148 & 8.884 & 8.262 & 8.262 & 12.635 & 11.674 & 11.226 & 13.612 & 13.711 \\
     \hline
    $p=2.0$ & 0.771 & 0.217 & 0.432 & 0.407 & 0.403 & 0.603 & 0.553 & 0.537 & 0.645 & 0.645 \\ 
     \hline
    \end{tabular}
    \caption{Median Difference as a function of $n$ and $p$ for $r=16$}
    \label{tab:Median Table r=16}
\end{table}

\begin{figure*}[H]
    \centering
\begin{subfigure}
        \centering
        \includegraphics[scale=0.32]{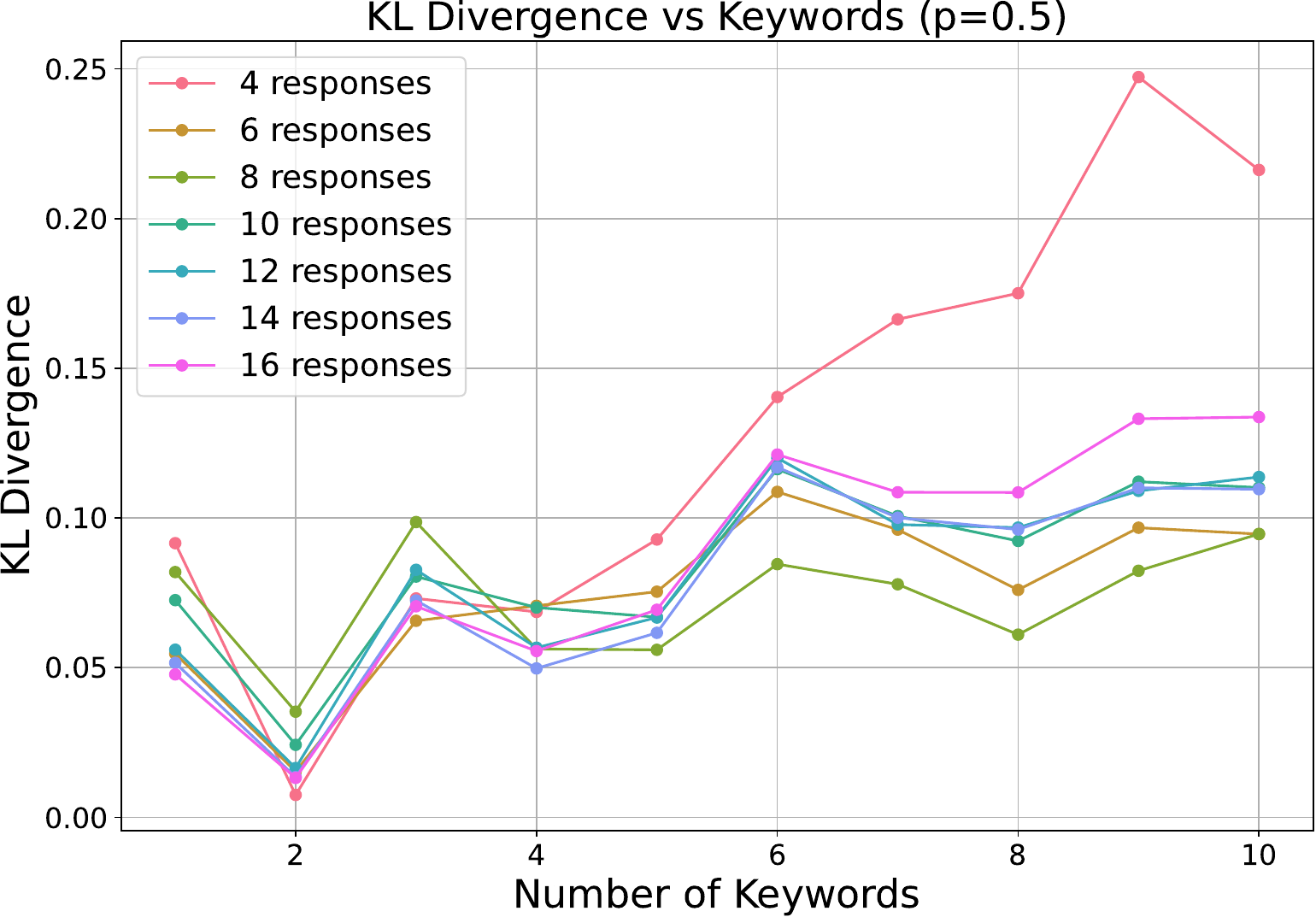}
    \end{subfigure}
    \hfill
\begin{subfigure}
        \centering
        \includegraphics[scale=0.32]{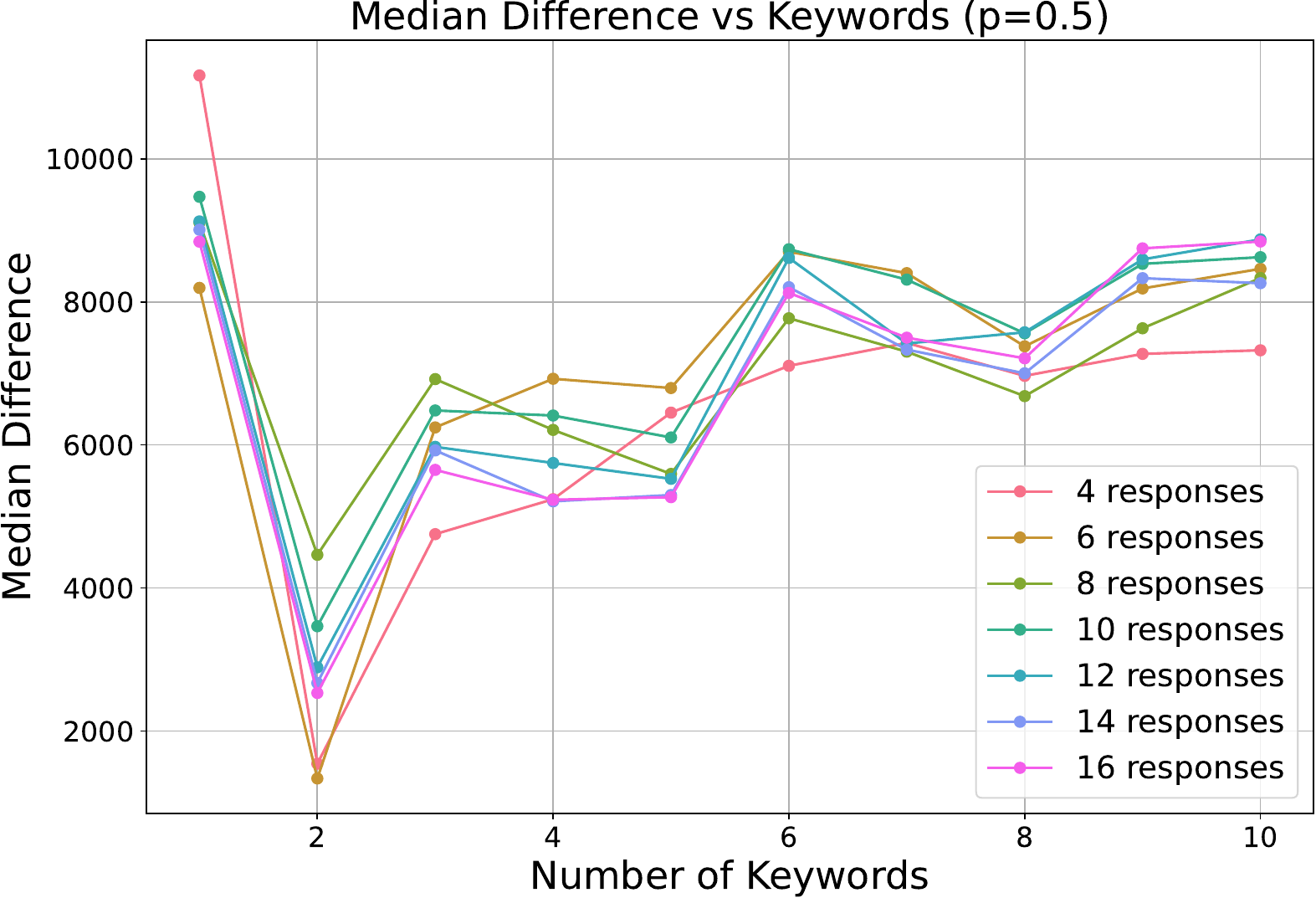}
    \end{subfigure}
    \caption{KL Divergence and Median Difference vs $n$ at different $r$ for $p=0.5$ using training data.}
    \label{fig:kldelta0.5}
\end{figure*}

\begin{figure*}[H]
    \centering
\begin{subfigure}
        \centering
        \includegraphics[scale=0.325]{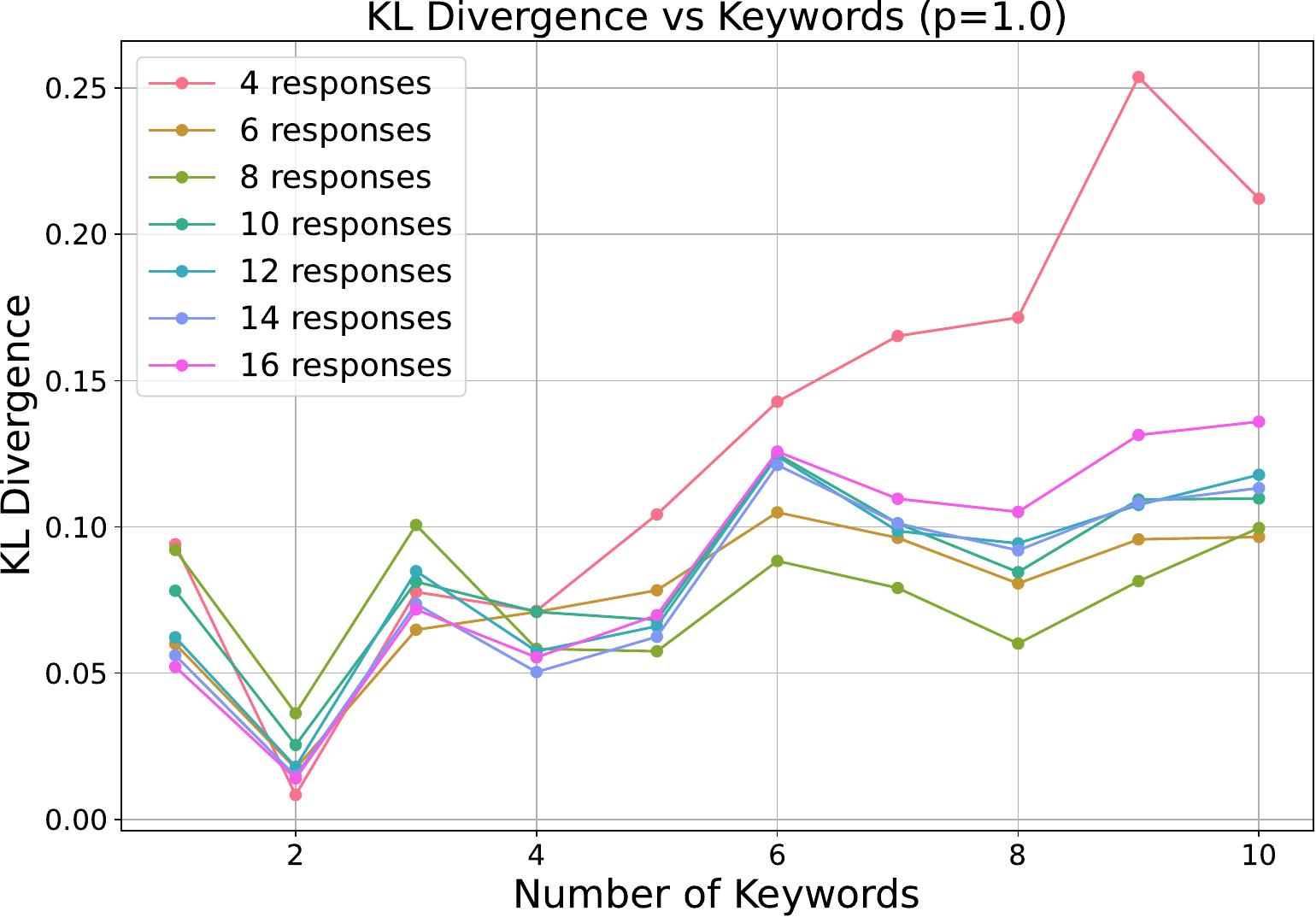}
    \end{subfigure}
    \hfill
\begin{subfigure}
        \centering
        \includegraphics[scale=0.325]{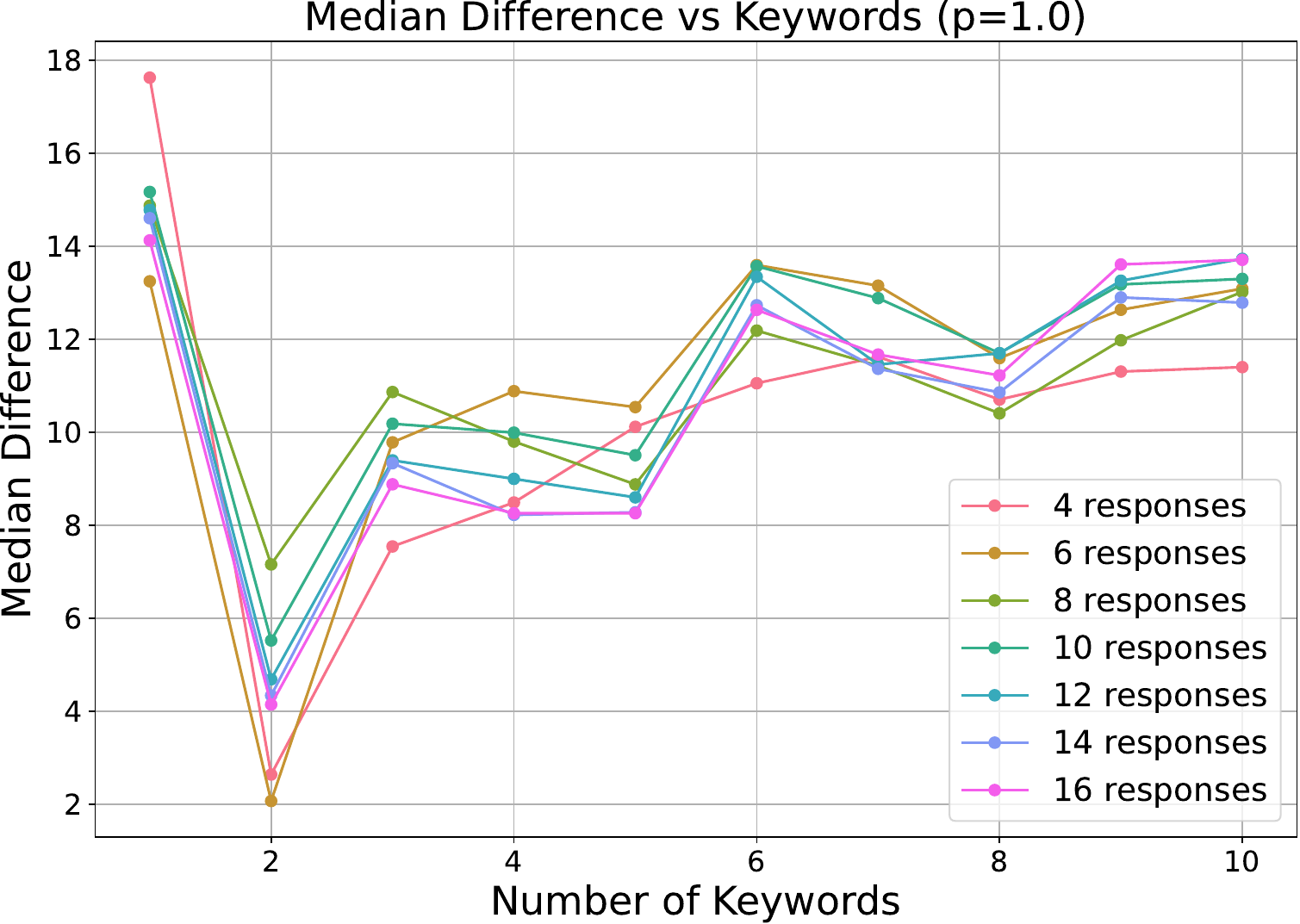}
    \end{subfigure}
    \caption{KL Divergence and Median Difference vs $n$ at different $r$ for $p=1.0$ using training data.}
    \label{fig:kldelta1.0}
\end{figure*}

\subsection{Test Results}
\subsubsection{$p=0.5$}

\begin{figure}[H]
    \centering
\begin{subfigure}
        \centering
        \includegraphics[width=0.49\columnwidth]{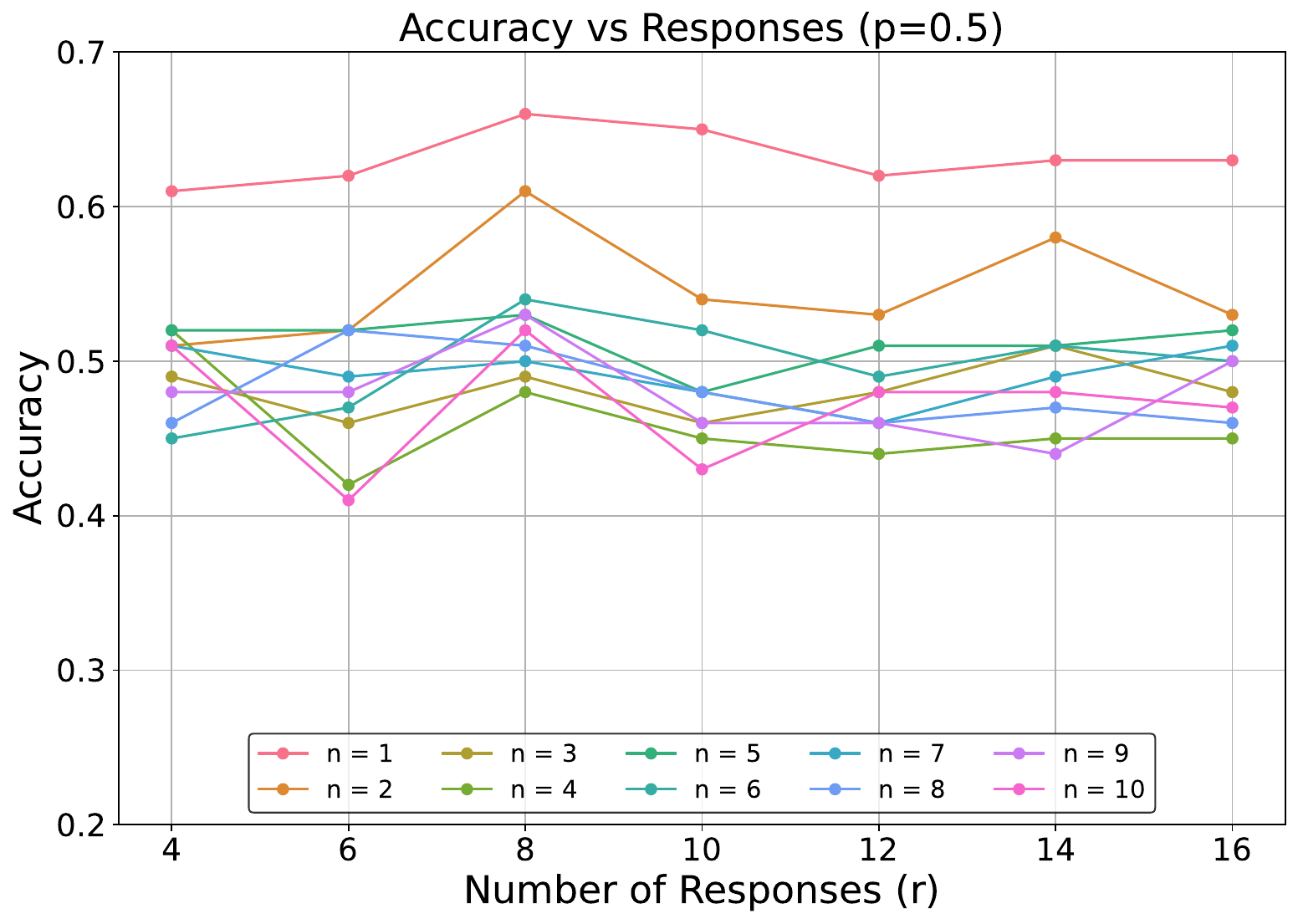}
    \end{subfigure}
    \hfill
\begin{subfigure}
        \centering
        \includegraphics[width=0.49\columnwidth]{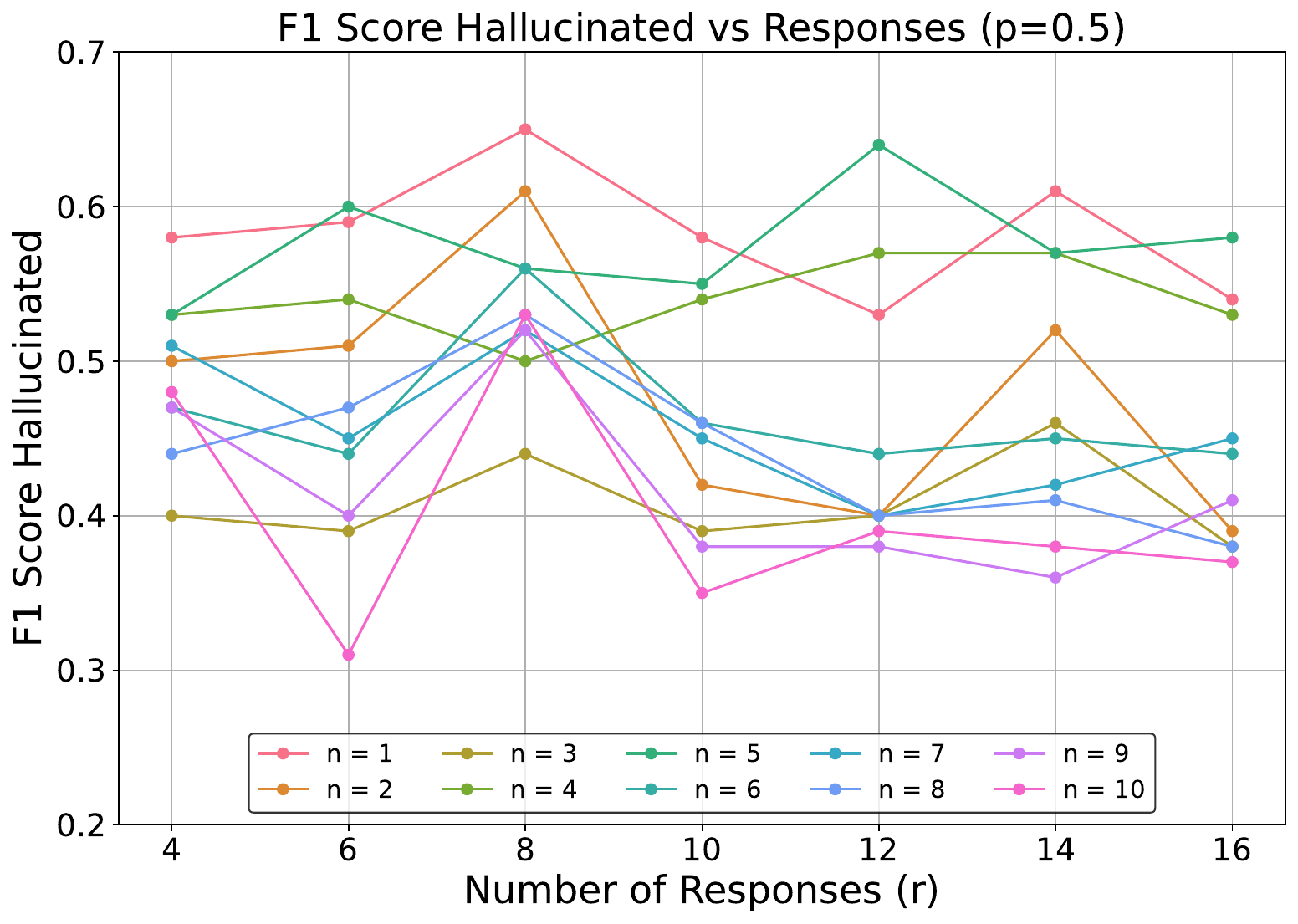}
    \end{subfigure}
    \caption{Model accuracy and hallucination F1 score vs $r$ at different $n$ for $p=0.5$}
    \label{fig:test0.5}
\end{figure}

\begin{minipage}[t]{0.48\columnwidth}
\begin{table}[H]
    \centering
    \small
    \setlength{\tabcolsep}{2pt}
    \begin{tabular}{||c|c|c|c|c|c|c|c||}
        \hline
        & $r=4$ & $r=6$ & $r=8$ & $r=10$ & $r=12$ & $r=14$ & $r=16$ \\ \hline\hline
        $n=1$ & 0.61 & 0.62 & 0.66 & 0.65 & 0.62 & 0.63 & 0.63 \\ \hline
        $n=2$ & 0.51 & 0.52 & 0.61 & 0.54 & 0.53 & 0.58 & 0.53 \\ \hline
        $n=3$ & 0.49 & 0.46 & 0.49 & 0.46 & 0.48 & 0.51 & 0.48 \\ \hline
        $n=4$ & 0.52 & 0.42 & 0.48 & 0.45 & 0.44 & 0.45 & 0.45 \\ \hline
        $n=5$ & 0.52 & 0.52 & 0.53 & 0.48 & 0.51 & 0.51 & 0.52 \\ \hline
        $n=6$ & 0.45 & 0.47 & 0.54 & 0.52 & 0.49 & 0.51 & 0.50 \\ \hline
        $n=7$ & 0.51 & 0.49 & 0.50 & 0.48 & 0.46 & 0.49 & 0.51 \\ \hline
        $n=8$ & 0.46 & 0.52 & 0.51 & 0.48 & 0.46 & 0.47 & 0.46 \\ \hline
        $n=9$ & 0.48 & 0.48 & 0.53 & 0.46 & 0.46 & 0.44 & 0.50 \\ \hline
        $n=10$ & 0.51 & 0.41 & 0.52 & 0.43 & 0.48 & 0.48 & 0.47 \\ \hline
    \end{tabular}
    \caption{Accuracy as a function of $r$ and $n$ for $p=0.5$}
    \label{tab:accuracy_p0.5}
\end{table}
\end{minipage}
\hfill
\begin{minipage}[t]{0.48\columnwidth}
\begin{table}[H]
    \centering
    \small
    \setlength{\tabcolsep}{2pt}
    \begin{tabular}{||c|c|c|c|c|c|c|c||}
        \hline
        & $r=4$ & $r=6$ & $r=8$ & $r=10$ & $r=12$ & $r=14$ & $r=16$ \\ \hline\hline
        $n=1$ & 0.58 & 0.59 & 0.65 & 0.58 & 0.53 & 0.61 & 0.54 \\ \hline
        $n=2$ & 0.50 & 0.51 & 0.61 & 0.42 & 0.40 & 0.52 & 0.39 \\ \hline
        $n=3$ & 0.40 & 0.39 & 0.44 & 0.39 & 0.40 & 0.46 & 0.38 \\ \hline
        $n=4$ & 0.53 & 0.54 & 0.50 & 0.54 & 0.57 & 0.57 & 0.53 \\ \hline
        $n=5$ & 0.53 & 0.60 & 0.56 & 0.55 & 0.64 & 0.57 & 0.58 \\ \hline
        $n=6$ & 0.47 & 0.44 & 0.56 & 0.46 & 0.44 & 0.45 & 0.44 \\ \hline
        $n=7$ & 0.51 & 0.45 & 0.52 & 0.45 & 0.40 & 0.42 & 0.45 \\ \hline
        $n=8$ & 0.44 & 0.47 & 0.53 & 0.46 & 0.40 & 0.41 & 0.38 \\ \hline
        $n=9$ & 0.47 & 0.40 & 0.52 & 0.38 & 0.38 & 0.36 & 0.41 \\ \hline
        $n=10$ & 0.48 & 0.31 & 0.53 & 0.35 & 0.39 & 0.38 & 0.37 \\ \hline
    \end{tabular}
    \caption{Hallucination F1 scores as a function of $r$ and $n$ for $p=0.5$}
    \label{tab:f1_scores_p0.5}
\end{table}
\end{minipage}

\subsubsection{$p=1.0$}
\begin{figure}[H]
    \centering
\begin{subfigure}
        \centering
        \includegraphics[width=0.49\columnwidth]{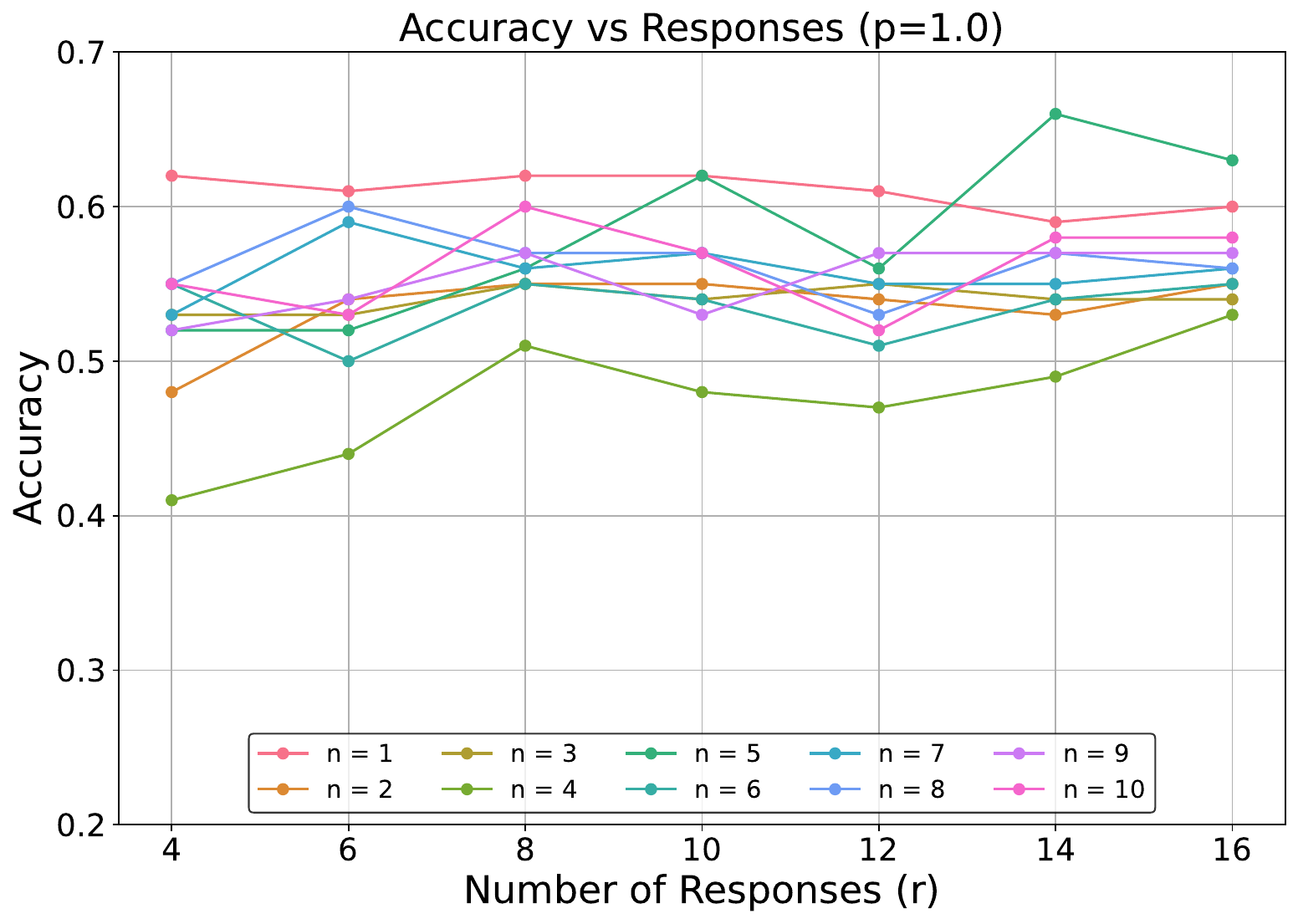}
    \end{subfigure}
    \hfill
\begin{subfigure}
        \centering
        \includegraphics[width=0.49\columnwidth]{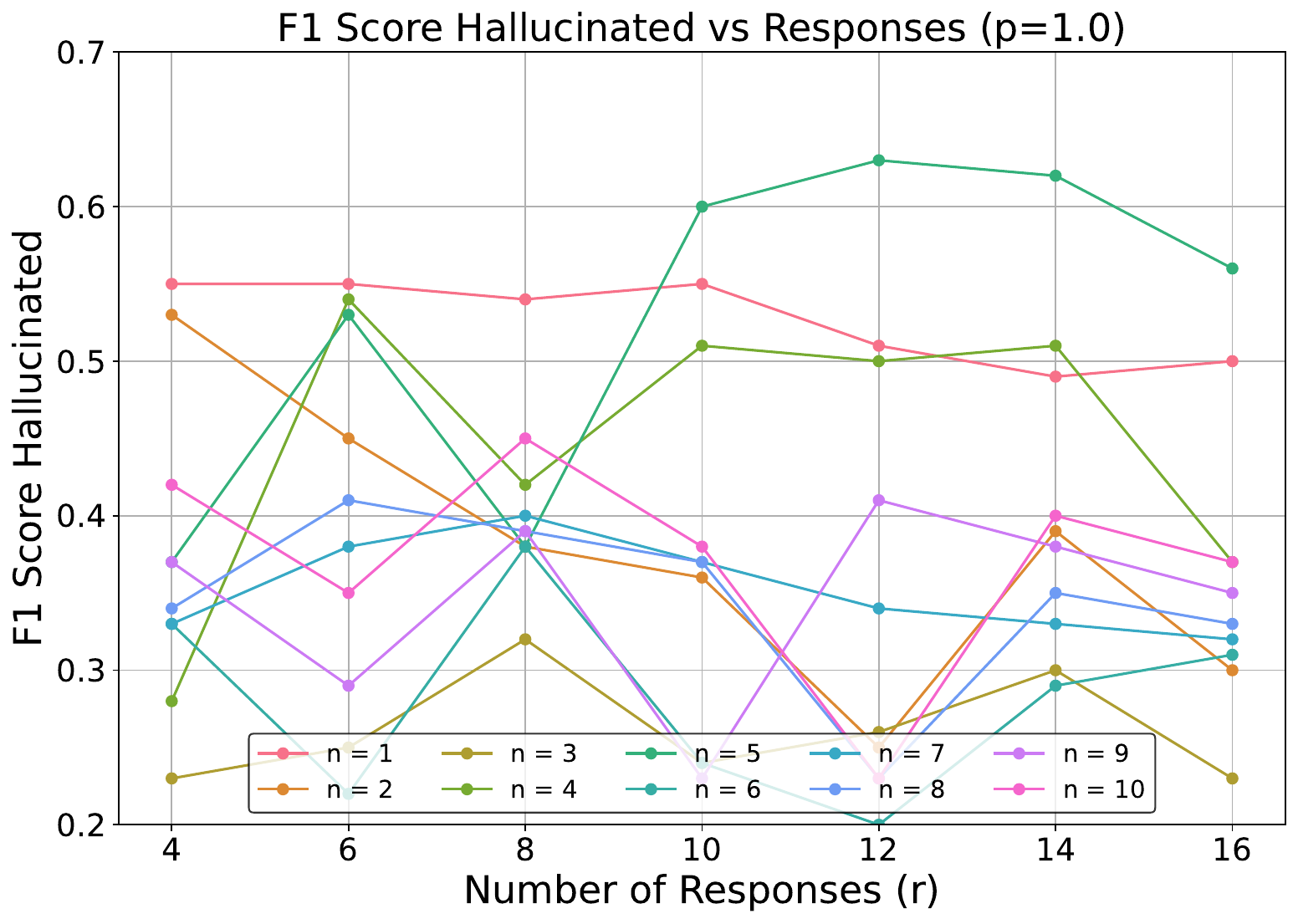}
    \end{subfigure}
    \caption{Model accuracy and hallucination F1 score vs $r$ at different $n$ for $p=1.0$}
    \label{fig:test1.0}
\end{figure}

\begin{minipage}[t]{0.48\columnwidth}
\begin{table}[H]
    \centering
    \small
    \setlength{\tabcolsep}{2pt}
    \begin{tabular}{||c|c|c|c|c|c|c|c||}\hline
        & $r=4$ & $r=6$ & $r=8$ & $r=10$ & $r=12$ & $r=14$ & $r=16$ \\ \hline
        $n=1$ & 0.62 & 0.61 & 0.62 & 0.62 & 0.61 & 0.59 & 0.6 \\ \hline
        $n=2$ & 0.48 & 0.54 & 0.55 & 0.55 & 0.54 & 0.53 & 0.55 \\ \hline
        $n=3$ & 0.53 & 0.53 & 0.55 & 0.54 & 0.55 & 0.54 & 0.54 \\ \hline
        $n=4$ & 0.41 & 0.44 & 0.51 & 0.48 & 0.47 & 0.49 & 0.53 \\ \hline
        $n=5$ & 0.52 & 0.52 & 0.56 & 0.62 & 0.56 & 0.66 & 0.63 \\ \hline
        $n=6$ & 0.55 & 0.5 & 0.55 & 0.54 & 0.51 & 0.54 & 0.55 \\ \hline
        $n=7$ & 0.53 & 0.59 & 0.56 & 0.57 & 0.55 & 0.55 & 0.56 \\ \hline
        $n=8$ & 0.55 & 0.6 & 0.57 & 0.57 & 0.53 & 0.57 & 0.56 \\ \hline
        $n=9$ & 0.52 & 0.54 & 0.57 & 0.53 & 0.57 & 0.57 & 0.57 \\ \hline
        $n=10$ & 0.55 & 0.53 & 0.6 & 0.57 & 0.52 & 0.58 & 0.58 \\ \hline
    \end{tabular}
    \caption{Accuracy as a function of $r$ and $n$ for $p=1.0$}
    \label{tab:accuracy_p1.0}
\end{table}
\end{minipage}
\hfill
\begin{minipage}[t]{0.48\columnwidth}
\begin{table}[H]
    \centering
    \small
    \setlength{\tabcolsep}{2pt}
    \begin{tabular}{||c|c|c|c|c|c|c|c||}
        \hline
        & $r=4$ & $r=6$ & $r=8$ & $r=10$ & $r=12$ & $r=14$ & $r=16$ \\ \hline\hline
        $n=1$ & 0.55 & 0.55 & 0.54 & 0.55 & 0.51 & 0.49 & 0.50 \\ \hline
        $n=2$ & 0.53 & 0.45 & 0.38 & 0.36 & 0.25 & 0.39 & 0.30 \\ \hline
        $n=3$ & 0.23 & 0.25 & 0.32 & 0.24 & 0.26 & 0.30 & 0.23 \\ \hline
        $n=4$ & 0.28 & 0.54 & 0.42 & 0.51 & 0.50 & 0.51 & 0.37 \\ \hline
        $n=5$ & 0.37 & 0.53 & 0.38 & 0.60 & 0.63 & 0.62 & 0.56 \\ \hline
        $n=6$ & 0.33 & 0.22 & 0.38 & 0.24 & 0.20 & 0.29 & 0.31 \\ \hline
        $n=7$ & 0.33 & 0.38 & 0.40 & 0.37 & 0.34 & 0.33 & 0.32 \\ \hline
        $n=8$ & 0.34 & 0.41 & 0.39 & 0.37 & 0.23 & 0.35 & 0.33 \\ \hline
        $n=9$ & 0.37 & 0.29 & 0.39 & 0.23 & 0.41 & 0.38 & 0.35 \\ \hline
        $n=10$ & 0.42 & 0.35 & 0.45 & 0.38 & 0.23 & 0.40 & 0.37 \\ \hline
    \end{tabular}
    \caption{Hallucination F1 scores as a function of $r$ and $n$ for $p=1.0$}
    \label{tab:f1_scores_p1.0}
\end{table}
\end{minipage}

\subsubsection{$p=2.0$}
The plot related to model accuracy and hallucination F1 score vs $r$ at different $n$ for $p = 2.0$ is already reported in the paper in Figure \ref{fig:test2.0}

\begin{minipage}[t]{0.48\columnwidth}
\begin{table}[H]
    \centering
    \small
    \setlength{\tabcolsep}{2pt}
    \begin{tabular}{||c|c|c|c|c|c|c|c||}
        \hline
        & $r=4$ & $r=6$ & $r=8$ & $r=10$ & $r=12$ & $r=14$ & $r=16$ \\ \hline\hline
        $n=1$ & 0.62 & 0.61 & 0.61 & 0.61 & 0.59 & 0.58 & 0.58 \\ \hline
        $n=2$ & 0.49 & 0.54 & 0.54 & 0.57 & 0.56 & 0.51 & 0.57 \\ \hline
        $n=3$ & 0.53 & 0.56 & 0.55 & 0.57 & 0.56 & 0.54 & 0.57 \\ \hline
        $n=4$ & 0.50 & 0.57 & 0.52 & 0.57 & 0.55 & 0.56 & 0.56 \\ \hline
        $n=5$ & 0.53 & 0.55 & 0.57 & 0.58 & 0.56 & 0.57 & 0.58 \\ \hline
        $n=6$ & 0.53 & 0.55 & 0.57 & 0.55 & 0.56 & 0.55 & 0.56 \\ \hline
        $n=7$ & 0.54 & 0.59 & 0.57 & 0.56 & 0.55 & 0.55 & 0.56 \\ \hline
        $n=8$ & 0.52 & 0.57 & 0.55 & 0.56 & 0.56 & 0.56 & 0.56 \\ \hline
        $n=9$ & 0.55 & 0.54 & 0.58 & 0.56 & 0.57 & 0.57 & 0.59 \\ \hline
        $n=10$ & 0.55 & 0.59 & 0.59 & 0.55 & 0.55 & 0.55 & 0.57 \\ \hline
    \end{tabular}
    \caption{Accuracy as a function of $r$ and $n$ for $p=2.0$}
    \label{tab:accuracy_p2.0}
\end{table}
\end{minipage}
\hfill
\begin{minipage}[t]{0.48\columnwidth}
\begin{table}[H]
    \centering
    \small
    \setlength{\tabcolsep}{2pt}
    \begin{tabular}{||c|c|c|c|c|c|c|c||}
        \hline
        & $r=4$ & $r=6$ & $r=8$ & $r=10$ & $r=12$ & $r=14$ & $r=16$ \\ \hline\hline
        $n=1$ & 0.56 & 0.55 & 0.53 & 0.52 & 0.50 & 0.49 & 0.48 \\ \hline
        $n=2$ & 0.54 & 0.40 & 0.37 & 0.42 & 0.35 & 0.38 & 0.34 \\ \hline
        $n=3$ & 0.21 & 0.36 & 0.33 & 0.35 & 0.33 & 0.33 & 0.34 \\ \hline
        $n=4$ & 0.30 & 0.54 & 0.40 & 0.44 & 0.37 & 0.37 & 0.36 \\ \hline
        $n=5$ & 0.35 & 0.42 & 0.37 & 0.39 & 0.40 & 0.36 & 0.36 \\ \hline
        $n=6$ & 0.29 & 0.42 & 0.39 & 0.33 & 0.35 & 0.32 & 0.33 \\ \hline
        $n=7$ & 0.31 & 0.46 & 0.38 & 0.35 & 0.35 & 0.35 & 0.34 \\ \hline
        $n=8$ & 0.25 & 0.41 & 0.33 & 0.34 & 0.34 & 0.33 & 0.34 \\ \hline
        $n=9$ & 0.31 & 0.35 & 0.40 & 0.35 & 0.39 & 0.38 & 0.39 \\ \hline
        $n=10$ & 0.36 & 0.49 & 0.42 & 0.37 & 0.38 & 0.38 & 0.39 \\ \hline
    \end{tabular}
    \caption{Hallucination F1 scores as a function of $r$ and $n$ for $p=2.0$}
    \label{tab:f1_scores_p2.0}
\end{table}
\end{minipage}

\end{document}